%% file: main.tex
\documentclass{article}

\usepackage{microtype}
\usepackage{graphicx}
\usepackage{subcaption}
\usepackage{booktabs} 

\usepackage{hyperref}

\usepackage[accepted]{icml2025}

\usepackage{amsmath}
\usepackage{amssymb}
\usepackage{mathtools}
\usepackage{amsthm}

\usepackage[capitalize,noabbrev]{cleveref}

\theoremstyle{plain}

\theoremstyle{definition}

\theoremstyle{remark}

\usepackage[textsize=tiny]{todonotes}

\usepackage{multirow}
\usepackage{makecell} 
\usepackage{bm} 
\usepackage{stmaryrd} 
\usepackage{dsfont} 
\usepackage{pifont} 
\usepackage{ifthen} 
\usepackage{booktabs} 
\usepackage{soul}
\usepackage{wrapfig} 
\usepackage{colortbl} 
\usepackage{xspace}

\usepackage{listings}
\usepackage{xcolor}

\definecolor{codegreen}{rgb}{0,0.6,0}
\definecolor{codegray}{rgb}{0.5,0.5,0.5}
\definecolor{codepurple}{rgb}{0.58,0,0.82}
\definecolor{backcolour}{rgb}{0.95,0.95,0.92}

\lstdefinestyle{mystyle}{
    backgroundcolor=\color{backcolour},   
    commentstyle=\color{codegreen},
    keywordstyle=\color{magenta},
    numberstyle=\tiny\color{codegray},
    stringstyle=\color{codepurple},
    basicstyle=\ttfamily\footnotesize,
    breakatwhitespace=false,         
    breaklines=true,                 
    captionpos=b,                    
    keepspaces=true,                 
    numbers=left,                    
    numbersep=5pt,                  
    showspaces=false,                
    showstringspaces=false,
    showtabs=false,                  
    tabsize=2
}
\definecolor{mediumtealblue}{rgb}{0.0, 0.33, 0.71}
\definecolor{darkpastelgreen}{rgb}{0.01, 0.75, 0.24}
\definecolor{azure}{rgb}{0.0, 0.5, 1.0}

\definecolor{crimsonred}{rgb}{0.86, 0.08, 0.24}
\definecolor{firebrick}{rgb}{0.7, 0.13, 0.13}
\definecolor{carmine}{rgb}{0.59, 0.0, 0.09}
\definecolor{rosewood}{rgb}{0.4, 0.0, 0.04}
\definecolor{deepcherry}{rgb}{0.6, 0.13, 0.13}

\newcommand{\ours}{\textcolor{azure}{(ours)}}
\newcommand{\apref}[1]{\ref{#1}}

\newcommand{\inlinesection}[1]{\noindent{\textbf{#1}}}

\newcommand{\projecturl}{https://snap-research.github.io/hpdm}

\newcommand{\Regname}{Scale Equivariance\xspace}
\newcommand{\regname}{scale equivariance\xspace}
\newcommand{\regshortname}{SE\xspace}

\newcommand{\regchfname}{Chopping High Frequencies\xspace}
\newcommand{\regchfshortname}{CHF\xspace}

\newcommand{\Diffusability}{\textit{Diffusability}\xspace}
\newcommand{\diffusability}{\textit{diffusability}\xspace}

\newcommand{\expect}[2][]{
\ifthenelse{\equal{#1}{}}{
\mathbb{E}\left[#2\right]
}{
\underset{#1}{\mathbb{E}}\left[#2\right]
}}

\DeclareMathOperator{\zigzag}{zigzag}

\DeclareMathOperator{\Dec}{Dec}

\newcommand{\cellbest}{\cellcolor{azure!35}}
\newcommand{\cellsecond}{\cellcolor{azure!10}}
\newcommand{\fid}{FID\xspace}
\newcommand{\fvd}{$\text{FVD}_\text{10K}$\xspace}
\newcommand{\fvdfivek}{$\text{FVD}_\text{5K}$\xspace}
\newcommand{\fvdfull}{$\text{FVD}_\text{50K}$\xspace}
\newcommand{\fidfivek}{$\text{FID}_\text{5K}$\xspace}
\newcommand{\fidtenk}{$\text{FID}_\text{10K}$\xspace}
\newcommand{\psnrsmall}{$\text{PSNR}_\text{512}$\xspace}
\newcommand{\dinofid}{$\text{FDD}$\xspace}
\newcommand{\dinofidfivek}{$\text{FDD}_\text{5K}$\xspace}

\newcommand{\cmsaei}{$\text{CMS-AE}_{I}$\xspace}
\newcommand{\cvae}{CV-AE\xspace}
\newcommand{\cvaefull}{CogVideoX-AE\xspace}
\newcommand{\dcae}{DC-AE\xspace}
\newcommand{\fluxae}{FluxAE\xspace}
\newcommand{\ltxae}{LTX-AE\xspace}

\newcommand{\loss}{\mathcal{L}}

\newcommand{\captionvspace}{\vspace{-1em}} 

\definecolor{DarkGreen}{RGB}{1,80,52}
\definecolor{DarkRed}{RGB}{150,30,30}

\newcommand{\papertitle}{Improving the Diffusability of Autoencoders}

\icmltitlerunning{\papertitle}

\begin{document}

\twocolumn[

\icmltitle{\papertitle}



\icmlsetsymbol{equal}{*}

\begin{icmlauthorlist}
\icmlauthor{Ivan Skorokhodov}{snap}
\icmlauthor{Sharath Girish}{snap}
\icmlauthor{Benran Hu}{snap,cmu}
\icmlauthor{Willi Menapace}{snap}
\icmlauthor{Yanyu Li}{snap}
\icmlauthor{Rameen Abdal}{snap}
\icmlauthor{Sergey Tulyakov}{snap}
\icmlauthor{Aliaksandr Siarohin}{snap}
\end{icmlauthorlist}

\icmlaffiliation{snap}{Snap Inc.}
\icmlaffiliation{cmu}{Carnegie Mellon University}

\icmlcorrespondingauthor{Ivan Skorokhodov}{iskorokhodov@gmail.com}
\icmlcorrespondingauthor{Aliaksandr Siarohin}{aliaksandr.siarohin@gmail.com}

\icmlkeywords{Diffusion Models, Latent Diffusion, Video Generation, Image Generation, AutoEncoders, VAE, Variational AutoEncoders, ICML}

\vskip 0.3in
]



\printAffiliationsAndNotice{}  

\input{sections/0-abstract}
\input{sections/1-introduction}
\input{sections/2-related-work}

\input{sections/3-method}

\input{sections/4-experiments}

\input{sections/5-conclusion}
\input{sections/6-impact}

\bibliography{main}
\bibliographystyle{icml2025}

\newpage
\appendix
\onecolumn

\include{supp/1-limitations}
\include{supp/2-details}
\include{supp/3-extra-exps}
\include{supp/4-visualizations}

\include{supp/5-failed-experiments}

\end{document}

%% file: sections/0-abstract.tex
\begin{abstract}
Latent diffusion models have emerged as the leading approach for generating high-quality images and videos, utilizing compressed latent representations to reduce the computational burden of the diffusion process.
While recent advancements have primarily focused on scaling diffusion backbones and improving autoencoder reconstruction quality, the interaction between these components has received comparatively less attention.
In this work, we perform a spectral analysis of modern autoencoders and identify inordinate high-frequency components in their latent spaces, which are especially pronounced in the autoencoders with a large bottleneck channel size.
We hypothesize that this high-frequency component interferes with the coarse-to-fine nature of the diffusion synthesis process and hinders the generation quality.
To mitigate the issue, we propose \emph{\regname}: a simple regularization strategy that aligns latent and RGB spaces across frequencies by enforcing scale equivariance in the decoder.
It requires minimal code changes and only up to $20$K autoencoder fine-tuning steps, yet significantly improves generation quality, reducing FID by 19\% for image generation on ImageNet-1K $256^2$ and FVD by at least $44\%$ for video generation on Kinetics-700 $17 \times 256^2$.
The source code is available at \url{https://github.com/snap-research/diffusability}.
\end{abstract}

%% file: sections/1-introduction.tex
\section{Introduction}
\label{sec:intro}

\input{figures/convergence}

In recent years, diffusion models (DMs) have emerged as the dominant generative modeling paradigm in computer vision.
However, the high dimensionality of visual data poses a significant challenge, making the direct application of diffusion models impractical.
Latent diffusion models (LDMs)~\cite{LSGM,LDM} have become the main approach in mitigating this issue, demonstrating remarkable success in generating high-resolution images~\cite{Flux, DALLE-3, SD3} and videos~\cite{Sora, CogVideoX, HunyuanVideo}.
A typical LDM consists of two main components: an autoencoder and a diffusion backbone.
Most recent breakthroughs have been driven by scaling up diffusion backbones~\cite{DiT}, while autoencoders (AEs) have received comparatively less attention.

Recently, the research community has begun focusing more on improving the autoencoders, recognizing their crucial impact on overall performance, but most effort has been concentrated on enhancing reconstruction quality~\cite{Flux, CogVideo, CosmosTokenizer, LTX-video, DC-AE} and achieving higher compression ratios~\cite{CosmosTokenizer, LTX-video, DC-AE} to accelerate the diffusion process.
However, we argue that a critical yet under-explored aspect, which we refer to as \diffusability\footnote{\Diffusability describes how easily a distribution can be modeled by diffusion: high \diffusability indicates that the distribution is easy to fit, while low diffusability implies 
a higher difficulty.}, also plays a key role in determining the utility of autoencoders.
Indeed, all three factors---reconstruction quality, compression efficiency, and diffusability---are essential for the practical effectiveness of LDMs.
Specifically, inaccurate reconstruction sets an upper bound on generation fidelity, low compression efficiency leads to slow and costly generation, and poor diffusability necessitates the use of heavier, more expensive, and sophisticated diffusion backbones, further limiting LDM quality.

Diffusion models possess a unique property of being coarse-to-fine in nature~\cite{spectral-autoregression, DCTdiff}: in the denoising process, they synthesize low-frequency signal components first and add high-frequency ones on top of them later.
It is a beneficial trait since it allows to defer error accumulation to higher frequency parts of the spectrum, which aligns well with how humans perceive quality: we are sensitive to the image structure and composition, but oblivious of its fine-grained textural details.
However, when applying the diffusion process in the latent spaces of pre-trained autoencoders, the correspondence between latent low-frequency components and their RGB counterparts may be lost, hindering the spectral autoregression property.

In this work, we identify a correlation between the spectral properties of the latent space and its diffusability.
We analyze the spectral characteristics of latent representations across several widely used image and video autoencoders.
Our investigation reveals a prominent high-frequency component in these latent spaces, deviating significantly from the spectral distribution of RGB signals.
This component becomes even more pronounced as the channel size increases, which the recent autoencoders use to improve reconstruction.
We hypothesize that the flat spectral distribution induced by the strong high-frequency component harms the spectral autoregression property.
Moreover, we demonstrate that these high-frequency components substantially influence the final RGB result, and their inaccurate modeling can introduce noticeable visual artifacts.
Finally, we show that standard KL regularization is insufficient to address spectrum defects and, in some cases, may even amplify the issue.

To mitigate these spurious high-frequency components in latent representations, we propose a simple and effective regularization strategy.
Our approach involves aligning the latent space and the RGB space at different frequencies. 
This is achieved by enforcing scale equivariance in the decoder — ensuring that downsampled latents correspond to downsampled RGB representations.
Our method requires minimal modifications and few additional autoencoder fine-tuning steps, yet significantly enhances diffusability across various architectures, ultimately improving the quality of generated samples.
We validate our approach on both image and video autoencoders, including FluxAE~\cite{Flux}, CosmosTokenizer~\cite{CosmosTokenizer}, CogVideoX-AE~\cite{CogVideo}, and LTX-AE~\cite{LTX-video}, consistently demonstrating improved LDM performance on ImageNet-1K~\cite{ImageNet} $256^2$, reducing FID by 19\% for DiT-XL, and Kinetics-700~\cite{kinetics700} $17\times 256^2$, reducing FVD by at least $44\%$.


%% file: figures/convergence.tex
\begin{figure}[ht]
\centering
\includegraphics[width=\linewidth,height=5cm]{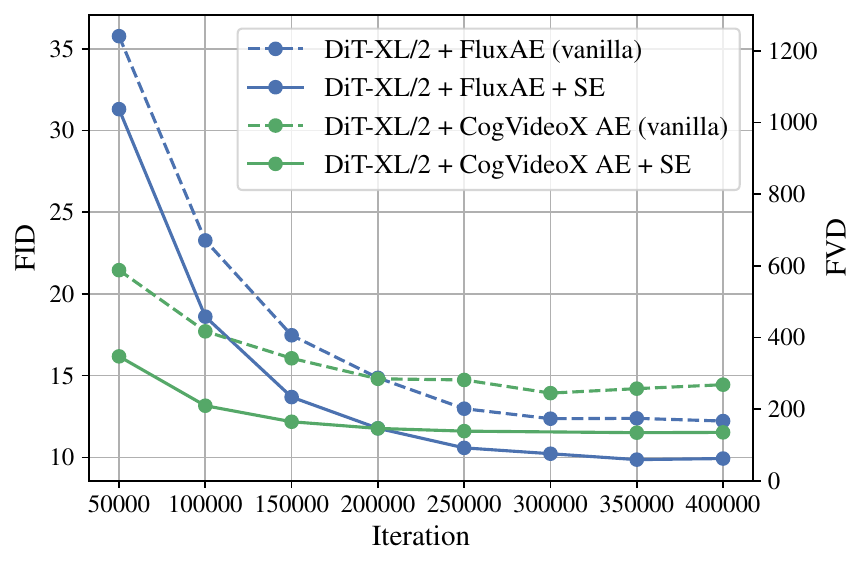}
\vspace{-2em}
\caption{Convergence speed of DiT-XL/2 on top of vanilla FluxAE vs FluxAE fine-tuned for 10K steps with \regname (\regshortname) regularization on ImageNet-1K-$256^2$; and on top of \cvaefull vs \cvaefull with \regshortname on Kinetics-700-$17\times256^2$. Our regularization improves the performance of image and video LDMs by refinng the frequency profile of their autoencoders' latent spaces.}
\label{fig:convergence}
\vspace{-1em}
\end{figure}

%% file: sections/2-related-work.tex
\section{Related Work}
\label{sec:related-work}

\inlinesection{Diffusion models.}
Diffusion models~\cite{LSGM,LDM,song_score_based,DDPM,IDDPM,EDM} have emerged as the dominant framework for generative modeling, surpassing traditional approaches like GANs~\cite{GANs,StyleGAN} and VAEs~\cite{VAE}. Diffusion models express generation as a denoising process producing the generated content by progressively denoising an initial noise sample. Owing to their efficiency and scalability,
foundational generative models~\cite{Imagen,ImagenVideo,CogVideoX,SDXL,SVD,MovieGen} have made significant strides in synthesizing visually stunning and semantically aligned images and videos.

Initially applied to low-resolution visual content in the pixel space~\cite{LSGM,DDPM,IDDPM,EDM}, they have soon been extended to higher resolutions. In Latent Diffusion Models (LDMs)~\cite{LSGM,LDM} high-resolution visual content is modeled in the compact latent space produced by a variational autoencoder (VAE)~\cite{VAE} within a two-stage framework. Latent Flow Models (LFMs)~\cite{LFM,Instaflow}, follow the same approach but leverage Rectified Flows (RFs) to enable faster and more stable sampling.


Recent work attributes the success of diffusion models to a form of implicit spectral autoregression~\cite{InverseHeatDissipation,DCTdiff} implied by the progressive removal of noise during sampling, resulting in the generation of visual content in a coarse-to-fine manner.
Such result holds in the pixel-space of natural images, based on its pattern of decreasing spectral power~\cite{ruderman1977}.
We show that popular autoencoders~\cite{Flux,CosmosTokenizer,LTX-video,CogVideoX} have a less pronounced pattern of decreasing spectral power, inhibiting implicit spectral autoregressive generation.
Building on this observation, this work proposes a regularization scheme that re-establishes this property, consistently showing improved LDM performance and avoiding the need for explicit coarse-to-fine generation.





\inlinesection{Image and video autoencoders}.
Due to the success of LDMs, a lot of effort has been devoted to the development of better AEs. 
Image LDMs \cite{LDM} and early video diffusion models \cite{SVD,AnimateDiff} employ a spatial AE with a compression ratio of $1{\times}8{\times}8$. 
The rapid advancement of video diffusion models poses the demand for 3D AEs that jointly compress spatial and temporal dimensions to further improve efficiency \cite{CogVideo,zhou2024allegro,HunyuanVideo}. 
Among them, Open-Sora \cite{opensora} inherits the $1{\times}8{\times}8$ spatial AE and trains a decoupled $4{\times}$ temporal AE on top of its latent space, while others tend to build a hierarchical spatio-temporal AE with 3D causal convolutions \cite{xing2024videovaeplus,wu2024ivvae,chen2024deep,zhao2024cvvae,sadat2024litevae,hansen2025vitok}. 
To accelerate the model and improve the reconstruction, Open-Sora-Plan~\cite{opensoraplan} and CosmosTokenizer~\cite{CosmosTokenizer} propose to employ the wavelet transform of the input. 
Another popular trend is to further increase the compression ratio to reduce the number of tokens in the latent space \cite{xie2024sana,tian2024reducio,LTX-video}, thus enabling a more efficient denoising process. 
In addition to the continuous AEs explored in this work, multiple discrete AEs \cite{wang2024omnitokenizer,tang2024vidtok,CosmosTokenizer} are proposed to aid autoregressive tasks. \citet{esteves2024spectral} leverages a wavelet transform to produce latents corresponding to different frequency components. 

\inlinesection{AEs for compression}.
Many works train neural-based AEs for image compression~\cite{balle2016end,balle2018variational,minnen2018joint,cheng2020learned}, typically with $16{\times}16$ downsampled latents which are discrete and entropy constrained.
Video compression AEs involve autoregressive AEs~\cite{DCVC,DCVC-DC,DCVC-TCM} with explicit frame-wise 
formulations that utilize motion vectors or implicit modeling~\cite{VCT}.
These approaches target high-quality reconstruction with low bitrates and adopt complex designs for learnable entropy models.
They typically employ a larger number of latent bottleneck channels ($96{-}192$), which is not generally suited for the generation task, thus we do not consider them in this work.

\inlinesection{Concurrent works}.
Independently from us, AF-LDM~\cite{AliasFreeLDM} enforces shift equivariance in both the autoencoder and LDM, and EQ-VAE~\cite{EQVAE} proposes scale/shift equivariance regularization for autoencoders, but with a different motivation of improving the models' equivariance to spatial transformations.

%% file: sections/3-method.tex
\input{figures/spectrums-comparison-flux}
\input{figures/flux-kl-spectrums}

\section{Improving Diffusability}
\label{sec:method}

We begin this section by discussing the spectral decomposition of 2D signals and providing some background on discrete cosine transform in Section~\ref{sec:method:dct}.
In Section~\ref{sec:method:spec}, we analyze the spectral properties of latent spaces across different autoencoders and compare them to those of the RGB space.
Our main insight is that the frequency profile of the latent space includes large-magnitude high-frequency components.
We also show that as the channel size increases, the high-frequency components become more pronounced.
Additionally, we demonstrate that the widely adopted KL regularization only increases the strength of these components.
Finally, \cref{sec:method:scalereg} presents a straightforward method to improve the diffusability of a latent space of an autoencoder by enhancing its spectral properties.

\subsection{Background: Blockwise 2D DCT}
\label{sec:method:dct}


The discrete cosine transform (DCT)~\cite{ahmed2006discrete} over a 2D signal is a transformation converting the signal's representation between the spatial and frequency domains. DCT, in particular, represents the original input signal as coefficients for a set of horizontal and vertical cosine basis oscillating with different frequencies. 
More formally, given a 2D signal block $\mathbf{A} \in \mathbb{R}^{B \times B}$ whose values $P_{xy}$ denote the pixel intensity at position $(x,y)$, 
the two-dimensional type-II DCT yields a frequency-domain block $\mathbf{D} \in \mathbb{R}^{B \times B}$ where $D_{uv}$ captures the coefficient for the corresponding horizontal and vertical cosine bases:
\begin{align}
D_{uv} &= \alpha(u) \alpha(v) \sum_{x=0}^{B-1}\sum_{y=0}^{B-1} P_{xy} f(x, u) f(y, v), \nonumber \\ 
& \text{where} \quad \alpha(u) = \left\{\begin{matrix} 
\sqrt{1/B}, \quad u=0, \\
\sqrt{2/B}, \quad u\neq0,
\end{matrix}\right. \nonumber \\
& f(x, u) = \cos\left(\tfrac{(2x+1) u \pi}{2B}\right). \nonumber
\end{align}
In practice, we split the input 2D signal into non-overlapping blocks of size $B\times B$ and treat each channel independently.

By analyzing RGB images and latents in the DCT frequency domain, we produce a \emph{frequency profile} that relates to the energy of the signal at every frequency. 
A zigzag frequency index is used to map each DCT block $\mathbf{D}\in\mathbb{R}^{B\times B}$ into a one-dimensional sequence following the standard \emph{zigzag} ordering as in JPEG~\cite{JPEG}, which indexes the DCT coefficients from lowest frequency $D_{0,0}$ to highest one $D_{B-1,B-1}$.
Formally, let $\zigzag(u,v) \in \{0,\dots,B^2-1\}$ denote the ranks of the coefficients $D_{uv}$ in the ascending frequency order.
Given a block, we compute its DCT and produce the normalized amplitudes for each frequency component \((u, v)\) as:
\begin{equation}  
    A_{uv} = \left |\frac{D_{uv}}{D_{0,0}} \right|.  
\end{equation}  
We define the \emph{frequency profile} as the sequence of normalized amplitudes in the standard \emph{zigzag} order.

When analyzing the frequency profiles of videos (or latent codes with an additional time dimension), we still rely on per-frame 2D DCT since the temporal and spatial domains possess different spectral properties.

\subsection{Spectral Analysis of the Latent Space}
\label{sec:method:spec}

\input{figures/improved-spectrums}

We begin our analysis by observing the frequency profile of the latent space in the Flux~\cite{Flux} family of autoencoders to establish a relationship with \emph{diffusability}. 
For the purpose of this study, we train a family of FluxAE models with various channel sizes for 100k steps (where performance saturates in this setting) and, for each of them, compute the averaged frequency profile over 256 samples, all channels, and all DCT blocks. 
Figure~\ref{fig:spectrums-comparison-flux} presents the frequency profiles of both Flux autoencoders and RGB space, from which we observe: (i) The Flux profile exhibits significantly larger high-frequency components compared to the RGB profile. (ii) As the number of channels in the autoencoder's bottleneck increases, high-frequency components become more pronounced.
We hypothesize that a larger bottleneck allows the autoencoder to capture finer, high-frequency details. Initially, limited capacity prioritizes smoother, low-frequency information.
But as the capacity expands, the model encodes additional high-frequency content, distributing it unevenly and in the unstructured manner across channels.
This finding is of particular interest as the number of channels is positively correlated with autoencoder's reconstruction quality.

A common way to regularize the latent space for latent diffusion models (LDMs) is to employ a variational autoencoder (VAE)~\cite{VAE} framework with a KL divergence term, encouraging the latent distribution to align with the standard Gaussian prior.
Since the reverse diffusion process also starts with the same standard normal distribution, such KL regularization is argued to simplify the job for the diffusion model~\cite{LSGM}, since it now has less ``work'' to do.
However, as we show in Figure~\ref{fig:flux-kl-spectrums} which compares FluxAE with varying levels of KL regularization, higher KL regularization introduces more high frequencies due to the underlying noise addition process, creating a harmful side-effect, reducing diffusability.

Previous work~\cite{DCTdiff, spectral-autoregression, InverseHeatDissipation} interprets diffusion models as autoregressive ones, but in the spectral domain: when noise level is high, low frequencies are generated, then, as the level of noise lowers during sampling, progressively higher frequencies are generated.
This is an attractive property since it allows the model to leverage the cleaner lower frequencies as a conditioning signal for the current prediction.
However, the strength of this autoregressive pattern is directly related to the shape of the frequency profile for the signal to generate. Since the white noise that is applied as part of the diffusion process has a flat frequency profile, it follows that the flatter the frequency profile of the signal, the lower the cleanliness of low frequencies that can act as conditioning for the model. For a flat frequency profile, no autoregressive generation is possible as all frequencies would be erased at the same speed by white noise.
We also hypothesize that higher frequencies components are harder to model than lower frequency components for the following reasons and thus should be avoided: (i) they have higher dimensionality\footnote{E.g., going from $256^2$ to $512^2$ resolution adds little structural content, but increases the dimensionality $3$ times.}; (ii) they are generated only in the final steps of sampling, thus must emerge more rapidly; (iii) they are more susceptible to error accumulation over time~\cite{EPDM}.

Motivated by this analysis, we propose scale equivariance regularization for the autoencoder's latent space.


\input{figures/qualitative-dct-cut}

\subsection{\Regname Regularization}
\label{sec:method:scalereg}
Effective regularization should achieve two key objectives: (i) to suppress high-frequency components in the latent space and (ii) to prevent the decoder from amplifying these components, as their impact on the final result is what ultimately matters. This can be accomplished by aligning the spectral properties of the latent and RGB spaces at different frequencies.
A way to achieve this consists in explicitly chopping off a portion of the high frequencies in both spaces and training the decoder to reconstruct the truncated RGB signal from the truncated latent representation.
Our preliminary experiments demonstrated that an autoencoder can easily learn to alter its latent frequency profile to encode the inputs in the low-frequency region of the spectrum without sacrificing the reconstruction quality much (see \cref{fig:progressive-dct-cut}).
While this regularization, which we name \regchfname (\regchfshortname), improves the spectrum, we develop a much simpler procedure to achieve the same effect without the need to perform the error-prone DCT transform (the details of \regchfshortname are described in \cref{ap:freq-reg}).
The simplest way to achieve high-frequency truncation is through direct downsampling, which we discuss next.

Downsampling involves resizing both the input  $x$ and the latent representation $z$ by a fixed scale, yielding $\Tilde{x}$ and $\Tilde{z}$, respectively.
This process effectively removes a portion of the high-frequency components from both the RGB and latent signals.
In practice, we use $\times 2-4$ bilinear downsampling for all the experiments.
Regularization is then enforced by ensuring that $\Tilde{x}$ and the decoder’s reconstruction of the downsampled latent $\Dec(\Tilde{z})$ remain consistent through an additional reconstruction loss.
The autoencoder is trained using the following objective:  
\begin{equation}\label{eq:loss}
\loss(x) = d(x, \Dec(z)) + \alpha d( \Tilde{x}, \Dec(\Tilde{z}) ) + \beta\loss_\text{KL}.  
\end{equation}  

Here, $d(\cdot,\cdot)$ represents a distance measure for reconstruction which we instantiate as mean squared error loss and perceptual losses~\cite{lpips} following prior work~\cite{LDM}, $\alpha$ is the regularization strength (we use $\alpha = 0.25$ for the main experiments).
The term $\loss_\text{KL}$ is VAE's~\cite{VAE} KL regularization, if applicable (we do not use it when we train with our regularization).
This regularization effectively enforces scale equivariance in the decoder, which is the basis for its name.  

In Figure~\ref{fig:improved-spectrums}, we illustrate the effect of \regname on the spectrum of FluxAE. Our proposed regularization effectively reduces the high-frequency components of the signal, bringing it closer to the spectral characteristics of the RGB space. This successfully achieves objective (i) for effective regularization.
Meanwhile, Figure~\ref{fig:reg-dct-cut-flux} demonstrates that \regname preserves more content compared to the baseline, as more and more high-frequency components are suppressed, thereby fulfilling objective (ii).
Finally, Figure~\ref{fig:traj-comparison} visualizes intermediate steps in the diffusion trajectory. The regularized model exhibits a noticeably smoother and more structured progression, following a healthier coarse-to-fine generation process.
\vspace{-0.1cm}
\input{figures/traj-comparison}

%% file: figures/spectrums-comparison-flux.tex
\begin{figure}[t]
\centering
\includegraphics[width=\linewidth]{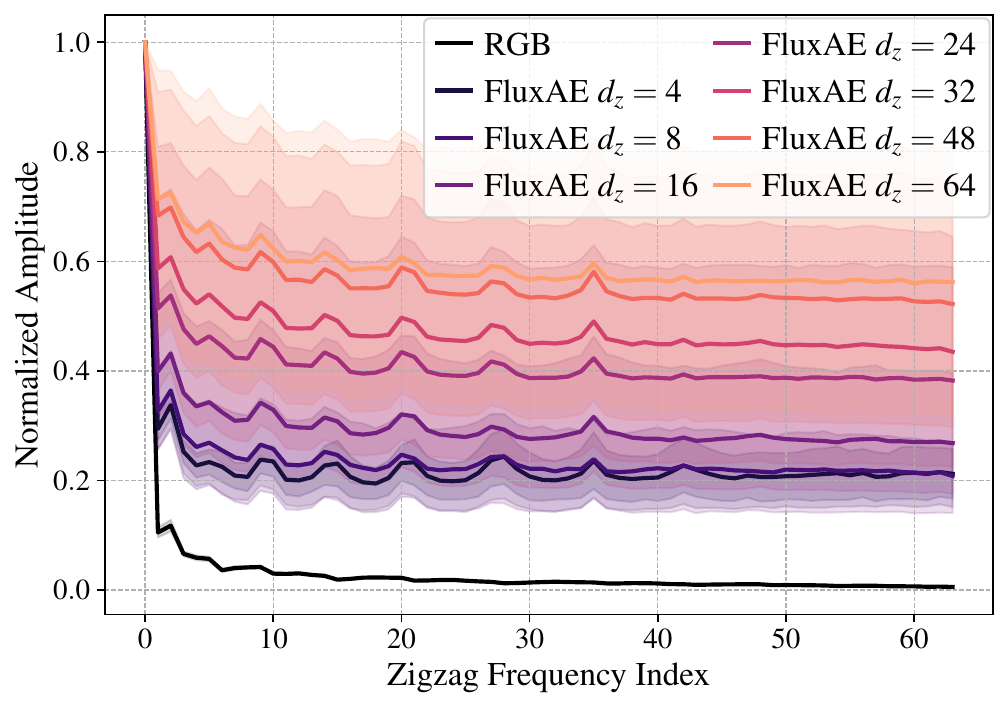}
\vspace{-2em}
\caption{Latent frequency profiles of FluxAE autoencoders of varying bottleneck sizes, and also RGB (of the same $32^2$ spatial dimension). One can notice two things: 1) the latent space of an autoencoder exhibits a different power profile from RGB; and 2) high frequency amplitudes increase with the latent channel size.} 
\label{fig:spectrums-comparison-flux}
\vspace{-1em}
\end{figure}

%% file: figures/flux-kl-spectrums.tex
\begin{figure}[t]
\centering
\includegraphics[width=0.95\linewidth]{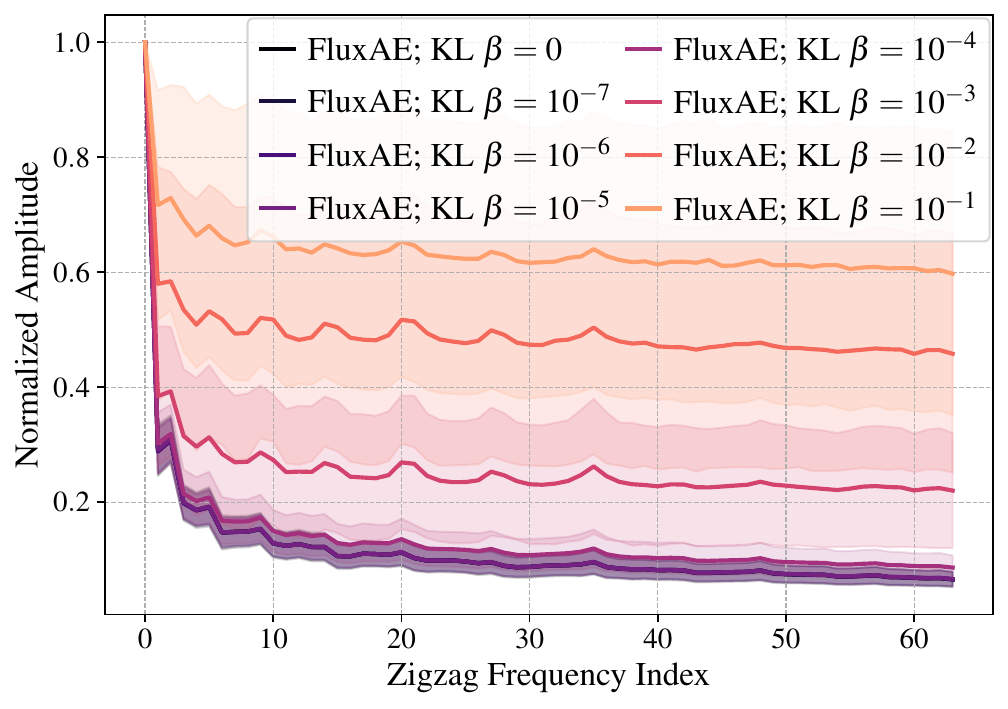}
\captionvspace
\caption{Spectrums for FluxAE autoencoders trained (from scratch) with different KL regularization strengths. KL regularization is a double-edged sword: it pushes the latents distribution closer to standard Gaussian (the distribution the reverse diffusion process starts with), so that the LDM has less work to do~\cite{LSGM}, but it also introduces high-frequency components into the latents due to the random noise addition (see \cref{fig:rgb-noise}), which LDM is forced to model as well.
\cref{fig:rgb-noise} shows the influence of noise addition on the frequency profile.
}
\label{fig:flux-kl-spectrums}
\vspace{-1em}
\end{figure}

%% file: figures/improved-spectrums.tex
\begin{figure}
\centering
\includegraphics[width=0.95\linewidth]{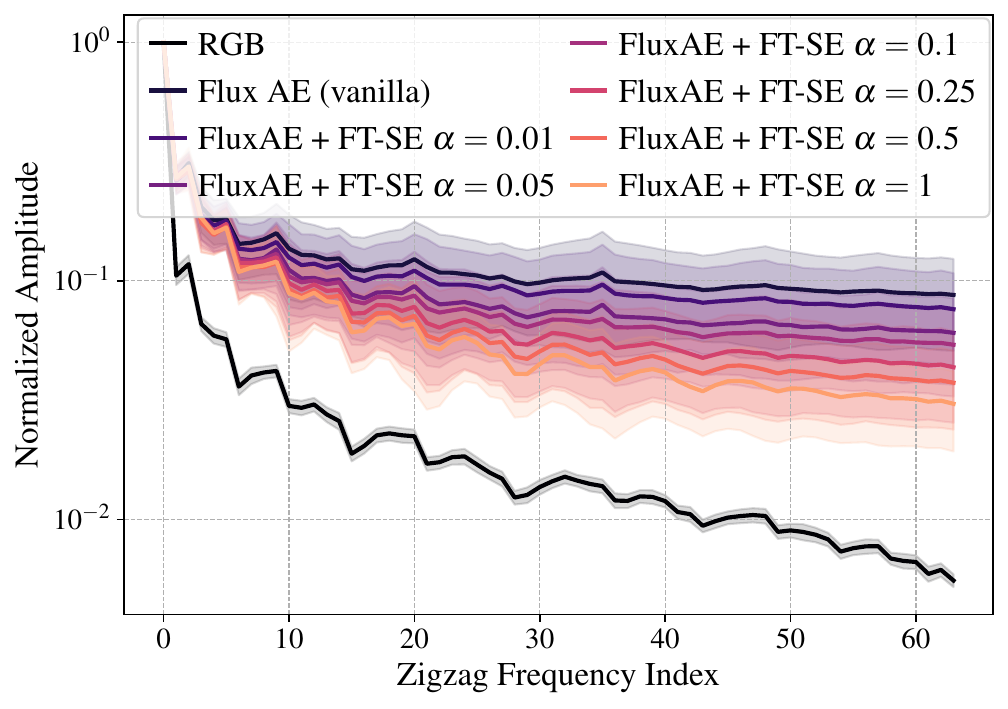}
\captionvspace
\caption{DCT Spectrum of the FluxAE latents with and without \regname (\regshortname) regularization. Fine-tuning AEs with \regshortname brings the spectrum closer to the RGB domain, the higher the regularization strength.}
\label{fig:improved-spectrums}
\vspace{-2em}
\end{figure}

%% file: figures/qualitative-dct-cut.tex
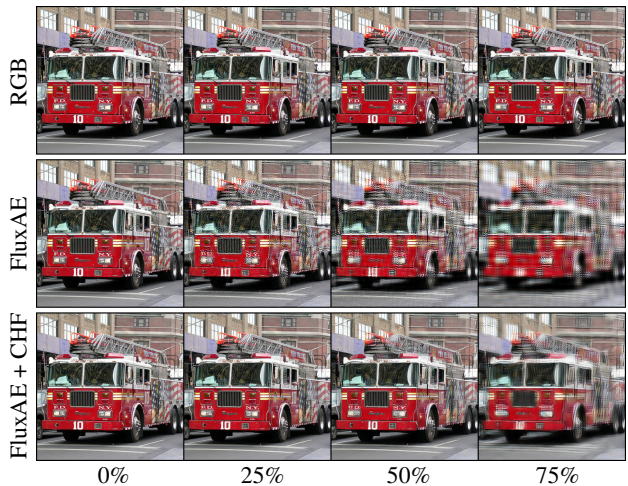
\begin{figure}[t]
\centering
\begin{minipage}{\linewidth}
\setlength{\unitlength}{\linewidth}
\centering
\vspace{-2.21cm}
\begin{picture}(1,1)
\put(0.05, 0){\includegraphics[width=0.95\linewidth]{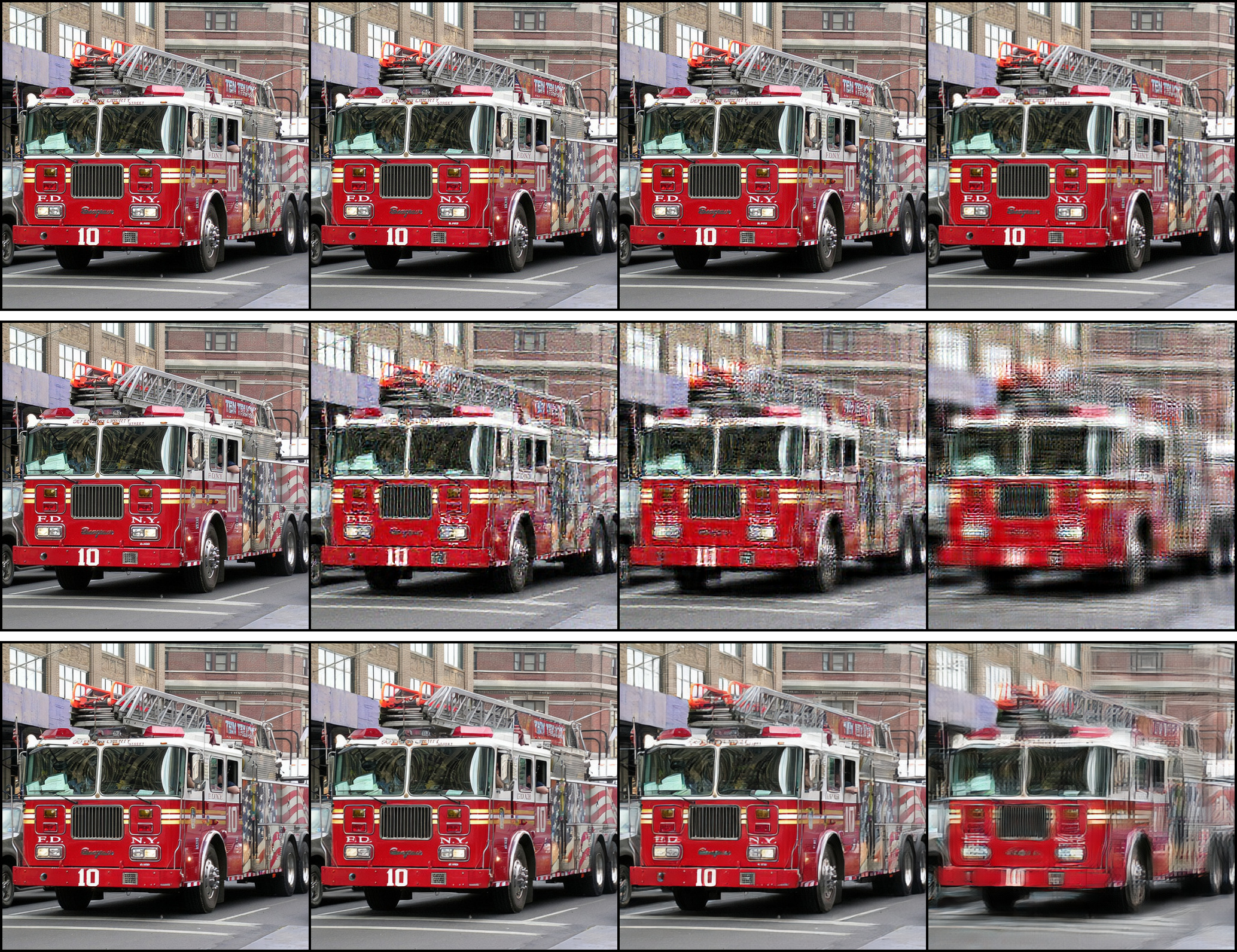}}

\put(0.175, -0.01){\makebox(0,0)[t]{\small 0\%}}
\put(0.415, -0.01){\makebox(0,0)[t]{\small 25\%}}
\put(0.65, -0.01){\makebox(0,0)[t]{\small 50\%}}
\put(0.89, -0.01){\makebox(0,0)[t]{\small 75\%}}

\put(0.035, 0.61){\makebox(0,0)[r]{\rotatebox{90}{\small RGB}}}
\put(0.035, 0.365){\makebox(0,0)[r]{\rotatebox{90}{\small FluxAE}}}
\put(0.035, 0.12){\makebox(0,0)[r]{\rotatebox{90}{\small FluxAE + CHF}}}
\end{picture}
\end{minipage}
\vspace{0em}
\caption{
RGB and autoencoder reconstructions with progressively erased DCT high-frequency components.
RGB faces minimal degradation (top), as a higher percentage of the latent DCT spectrum is removed, but the Flux AE reconstructions (middle) quickly degrade when the high-frequency components from the latents are being removed.
A high-frequency cutoff regularization forces the autoencoder to rely more on the low frequency region of the latents and leads to better compression and resilience to high-frequency error accumulation in diffusion models.
}
\label{fig:progressive-dct-cut}
\vspace{-1em}
\end{figure}

%% file: figures/traj-comparison.tex
\begin{figure}[t]
\centering
\includegraphics[width=\linewidth]{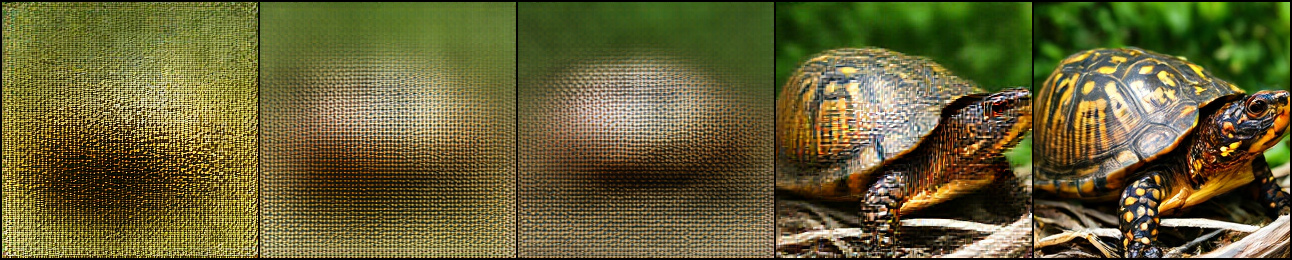}
\includegraphics[width=\linewidth]{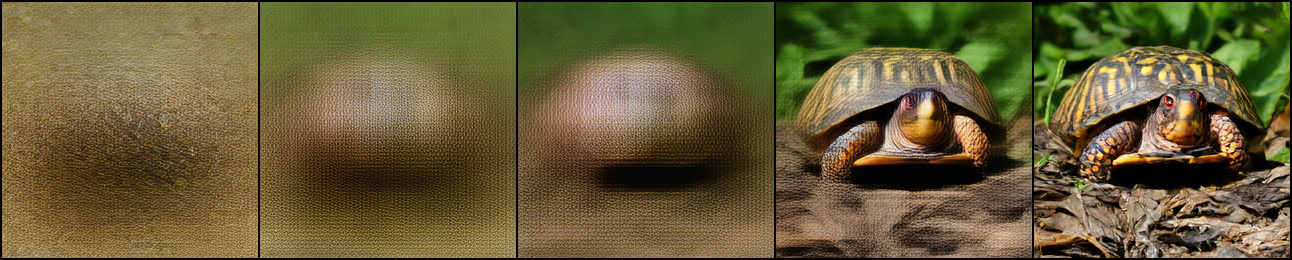}
\vspace{-2em}
\caption{Denoising trajectories (steps 1, 16, 32, 128 and 256 out of 256) for DiT-XL trained with FluxAE (top) and FluxAE+\regshortname. DiT-XL with vanilla FluxAE exhibits prominent high-frequency artifacts early on in the trajectory.}
\label{fig:traj-comparison}
\end{figure}

%% file: sections/4-experiments.tex
\section{Experiments}
\label{sec:experiments}

\input{figures/samples}
\input{tables/imagenet}

\inlinesection{Data.}
We trained all the autoencoders on in-the-wild data which do not overlap with ImageNet-1K~\cite{ImageNet} or Kinetics-700~\cite{kinetics700} to make sure that there is no data leak in the autoencoders, and that they remain general-purpose.
For this, we used our internal image and video datasets of the $256^2$ resolution, which are similar in distribution of concepts and aesthetics to the publicly available in-the-wild datasets like COYO~\cite{COYO} and Panda-70M~\cite{Panda70M}.
To control for the impact of the data (and also the training recipe), we trained a separate autoencoder baseline for each setup without using our proposed regularization.
In several cases, just fine-tuning on such in-the-wild data already yields better diffusion performance (e.g., see DiT-XL results in \cref{table:imagenet}).

\inlinesection{Evaluation.}
We evaluate image DiT models via \fid and \dinofid (Frechet Distance computed on top of DINOv2~\cite{DINOv2} features), where the latter was shown to be a more reliable metric~\cite{DinoFID, EDMv2}.
We evaluate video DiT models with \fvd~\cite{FVD}, \fid, and \dinofid, except for ablations where we rely on 5,000 samples.
For image models, we use 50,000 samples without any optimization for class balancing.
To evaluate autoencoders, we used PSNR, SSIM, LPIPS and \fid metrics computed on 512 samples from ImageNet and Kinetics-700 for image and video autoencoders, respectively.

\inlinesection{Training details.}
All the LDM models are trained for 400k steps with 10k warmup steps following the rectified flow diffusion parametrization~\cite{NormFlowsWithStochInterp, LFM, SD3}.
Following \citet{SD3}, we use a logit-normal training noise distribution.
We use either $2 \times 2$ or $1 \times 1$ patchification in DiT~\cite{DiT} to match the compute between DiTs trained on top autoencoders with different compression ratios.
Our video DiT is a direct adaption of the image one where we additionally unroll the temporal axis, following the recent works on video diffusion models~\cite{CogVideoX}.
We do not use patchification for the temporal axis in video DiTs.
In contrast to prior work (e.g., \citet{LDM}), we average the KL loss across the latent channels and resolutions: this has no theoretical impact, but allows to compare autoencoders with different bottleneck sizes.

\inlinesection{Inference details.}
We run DiT inference with 256 steps without classifier-free guidance~\cite{CFG} for quantitative evaluations since different models are too sensitive to it and should be tuned separately~\cite{EDMv2, CFG_in_interval}.


\subsection{Improving Existing Autoencoders}
\label{sec:experiments:main}
We apply our training pipeline on top of 3 different autoencoders.
For each autoencoder, we trained it while freezing the last output layers to avoid breaking their adversarial fine-tuning, which should have no impact on the latent space~\cite{DC-AE}.
We emphasize that \emph{none} of the explored modern autoencoders publicly released their training pipelines.
For the pretrained snapshots of autoencoders, we used the original snapshots available in the \verb|diffusers| library~\cite{diffusers}.

\inlinesection{Improving image autoencoders.}
For the image autoencoder, we used FluxAE~\cite{Flux} with $8 \times 8$ compression ratio and 16 latent channels (since it is the most popular modern autoencoder in the community) and \cmsaei~\cite{CosmosTokenizer} with $16 \times 16$ compression ratio and 16 channels as a high-compression autoencoder.
For all the experiments (unless stated otherwise), we fine-tuned it for just 10,000 training steps with a batch size of 32 (320,000 total seen images) using $2\times$ and $4\times$ downsampling ratios, chosen randomly during a forward pass of the regularization loss.
DiT training on top of the unchanged Flux Autoencoder is labeled as ``vanilla''.
Autoencoders fine-tuned for 10,000 steps with our proposed \regshortname regularization is denoted via the ``+ FT-\regshortname'' suffix.
To control for the fine-tuning data and training pipeline, we fine-tuned each autoencoder without adding our regularization as an additional loss (denoted via ``+ FT'').

For the LDM benchmark, we utilized ImageNet~\cite{ImageNet} at $256 \times 256$ resolution.
We used the DiT~\cite{DiT} model as the backbone since it is the most popular modern latent diffusion backbone.
Compared to the original paper, we incorporated several recent advancements into the DiT architecture to improve the baseline performance, as described in \cref{ap:details}.
Qualitative samples from DiT-XL/2 are provided in \cref{fig:samples}.
The results are shown in \cref{table:imagenet}.
One can observe that our proposed regularization greatly improves the \diffusability of the downstream LDM model, allowing to achieve 19\% lower \fid compared to the vanilla Flux AE and 8\% lower \fid compared to the Flux AE, fine-tuned in our training pipeline without the \regshortname regularization.

The improvement for \cmsaei is reduced with the main reason being that our training pipeline hurts its performance (we explored over 10 different hyperparameters setups to tune the vanilla model): after fine-tuning it for 10,000 steps with only reconstruction losses (it does not use KL regularization by default), the downstream \fid performance increases by 14\% from 11.69 to 13.59.

\inlinesection{Improving video autoencoders.}
For video autoencoders, we used \cvaefull~\cite{CogVideo} (\cvae) with $4 \times 8 \times 8$ compression and 16 latent channels and LTX-AE~\cite{LTX-video} with $8 \times 32 \times 32$ compression and 32 latent channels.
The latter serves as a strong high-compression autoencoder baseline.
All the video AEs are fine-tuned for 20,000 training steps on the joint image and video dataset with the batch size of 32.
Image batches are treated as single-frame videos which is possible due to the causal structure of the video autoencoders~\cite{MAGVITv2}.

\input{tables/kinetics}

\input{tables/kl-vs-downreg}

\input{figures/reg-dct-cut-flux}
\input{figures/ablations-downreg}

Similar to the image autoencoder experiments, we train a DiT model on Kinetics-700~\cite{kinetics700} on three variants: 1) a ``vanilla'' autoencoder snapshot; 2) the vanilla autoencoder fine-tuned for 20,000 steps in our training pipeline (denoted as ``FT''); and 3) the autoencoder snapshot, fine-tuned with our downsampling regularization (denoted as ``FT-\regshortname'').
For \ltxae, we used a reduced patchification resolution of $1 \times 1$ to compensate for its extreme compression ratio.
The results are presented in \cref{table:k700}: DiT model on top \regshortname-regularized autoencoders has drastically better performance: $44$\% and $54$\% lower \fvd for \cvae and \ltxae, respectively.
Our training pipeline allowed to achieve better DiT-B training for \cvae (650.4 vs 447.3 \fvd), but led to worse scores for \ltxae (854.4 vs 876.6), which we found less stable to train.
Adding our regularization strategy greatly improves the LDM performance in each case.
One can also observe that the boost for video autoencoders is larger than in the image domain (at least $-$44\% reduced \fvd for DiT-B for video generation vs $-7$\% reduced FID for DiT-XL for image generation).
We attribute this to two factors.
First, improvements in Frechet Distances~\cite{FID} do not scale linearly (i.e., their smaller values are progressively harder to improve).
Next, causal video autoencoders have less regular latent structure: the first frame in a video is encoded into the same representation size as the subsequent chunks, which leaves more room to enhance the diffusability of the latent space.

We additionally trained a DiT-XL/2 model for the \cvae family to explore the scalability of our regularization.
For this large-scale setup, our \regshortname regularization improved the \fvd score by almost twice.


\subsection{Ablations}
\label{sec:experiments:abl}
\input{tables/ae-rec}

\inlinesection{Does \regname hurt reconstruction?}
\regname in VAEs improves downstream generation quality in terms of FID (\cref{table:imagenet,table:k700}).
We now examine its impact on AE reconstruction quality. \Cref{table:ae-rec} presents results across four reconstruction metrics — PSNR, SSIM, LPIPS~\cite{lpips}, and FID, on 50,000 samples from ImageNet and Kinetics for image and video autoencoders, respectively. Reconstruction quality remains similar across the models. 

\inlinesection{Can LDM performance be improved by tweaking the KL weight instead?}
In \cref{table:kl-vs-downreg}, we show that increasing the KL strength can indeed improve the LDM performance for DiT-S/2, but it inevitably hurts reconstruction, as shown by the PSNRs, which bottlenecks the scalability of larger LDM models.
In contrast, our proposed \regname allows to achieve good LDM performance without hurting the reconstruction quality of the autoencoder.
For these ablations, we trained the DiT-S/2 variants for 200K training steps and DiT-L/2 variants for 400K steps, and ran inference for 80 steps without classifier-free guidance.
One can see that for a small compute budget, increasing the KL strength is beneficial: the best LDM score is obtained with the highest KL $\beta = 0.1$.
But it severely affected the reconstruction quality of the autoencoder, which limited its scaling: the corresponding DiT-L/2 LDM variant is ranked among the worst.
At the same time, our developed regularization performs well for all DiT variants and does limit scaling.

\inlinesection{Can the quality improvement come from implicit time shifting?}
\regshortname smoothes the latent space, which can result in ``implicit time shifting''~\cite{Lumina-T2X}: it eliminates high-frequency components from the latents, allowing the model to spend more compute on low-frequency ones, especially at inference time (see \apref{ap:details}).
We trained a family of DiT-B/2 models for various location and scale coefficients for logit-normal noise distribution~\cite{SD3}.
\cref{fig:time-shifting-abl} presents the $\text{FDD}_\text{5K}$ results with 128 inference steps (and without CFG) for time shifting coefficients of $[0.1, 0.25, 0.5, 0.75, 1, 2, 3, 4, 5, 6, 8, 10]$.
Our regularized FluxAE-FT+\regshortname achieves the best results across \emph{all} the setups, confirming that our improved quality is not due to improved diffusability, rather than implicit time shifting.
Note that our original noise schedule was tuned for the vanilla setup, which is why the default location of 0 and scale of 1 perform the best for non-regularized AEs.

\input{figures/ts-abl}

\inlinesection{Effect of \regshortname regularization strength.}
In \cref{fig:ablations-downreg}, we show the effect of varying the loss weight on the FluxAE performance on ImageNet.
Increasing the \regshortname regularization strength naturally worsens the total reconstruction quality since the decoder is trained to generalize its performance across both low frequency and high frequency latents (via downsampling) while having the same capacity. We choose the value of 0.25 to maintain reconstruction performance compared to the base AE model while improving the generation quality as shown in~\cref{table:imagenet,table:k700}.


\inlinesection{Effect of DCT spectrum cutting.} To examine the impact of downsampling regularization, we evaluate reconstruction quality by progressively removing high-frequency DCT components from the latents. \Cref{fig:reg-dct-cut-flux} presents reconstruction metrics on 512 samples from the ImageNet validation set for the baseline FluxAE and fine-tuning, with and without regularization. As the cut ratio increases on the x-axis, indicating the removal of more high-frequency components, the AE with regularized latents consistently achieves the best reconstruction quality across all metrics.
This underscores the regularization role in aligning the spectral properties of the latent and RGB spaces.

%% file: figures/samples.tex
\begin{figure*}
\centering
\includegraphics[width=\linewidth]{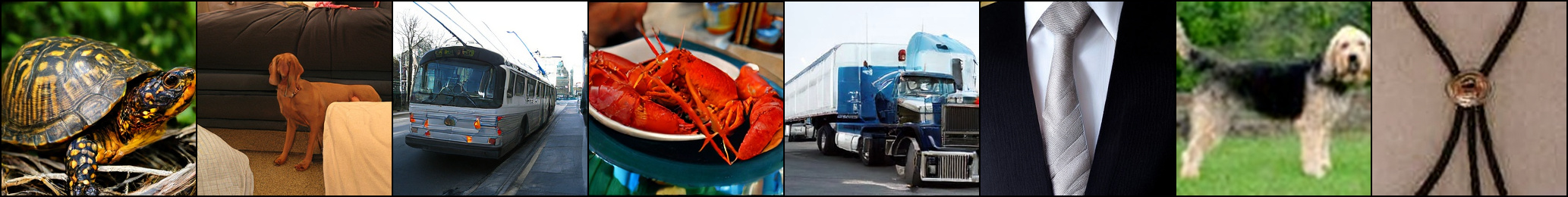}
\includegraphics[width=\linewidth]{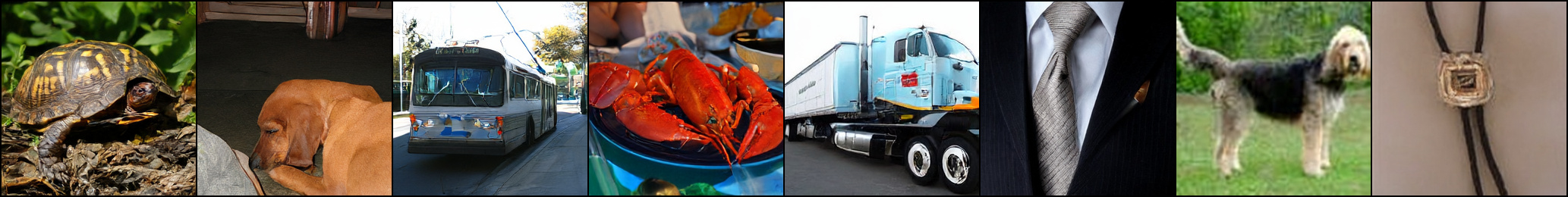}
\vspace{-2em}
\caption{Uncurated samples on ImageNet $256 \times 256$ from DiT-XL trained on top of FluxAE (top) vs DiT-XL with ``FluxAE + FT-\regshortname'' (bottom). 256 steps with the guidance scale of 3.0. More visualizations are in \cref{ap:visualizations}.}
\label{fig:samples}
\vspace{-2em}
\end{figure*}

%% file: tables/imagenet.tex
\begin{table}[ht]
\vspace{-1.5em}
\caption{Quantitative performance on ImageNet~\cite{ImageNet} without guidance. The original DiT scores are provided for reference from Table 4 of \cite{DiT}.}
\label{table:imagenet}
\centering
\resizebox{1.0\linewidth}{!}{
\begin{tabular}{llcc}
\toprule
Stage II & Stage I  & \fid$\downarrow$ & \dinofid$\downarrow$ \\
\midrule
\multirow{3}{*}{DiT-B/2} & \fluxae (vanilla) & \cellsecond{25.41} & \cellsecond{536.2} \\ 
& \fluxae + FT & 30.51 & 575.4 \\ 
& \fluxae + FT-\regshortname \ours & \cellbest{18.06} & \cellbest{450.6} \\ 
\midrule
\multirow{3}{*}{DiT-L/2} & \fluxae (vanilla) & \cellsecond{12.42} & \cellsecond{306.73} \\ 
& \fluxae + FT & 14.48 & 333.54 \\ 
& \fluxae + FT-\regshortname \ours & \cellbest{9.61} & \cellbest{236.43} \\ 
\midrule
\multirow{4}{*}{DiT-XL/2} & \fluxae (vanilla) & 12.21 & 282.8 \\
& \fluxae + FT & \cellsecond{10.62} & \cellsecond{262.2} \\
& \fluxae + FT-\regshortname \ours & \cellbest{9.85} & \cellbest{235.8} \\
&~ +$1$M steps & 3.27 & 85.86 \\
\midrule
\multirow{3}{*}{DiT-B/1} & \cmsaei (vanilla) & \cellbest{11.69} & \cellsecond{360.83} \\
& \cmsaei + FT & 13.59 & 375.19 \\
& \cmsaei + FT-\regshortname \ours & \cellsecond{11.85} & \cellbest{354.22} \\
\midrule
DiT-B/2 (orig) & \multirow{4}{*}{SD-VAE-ft-MSE} & 43.47 & $-$ \\
DiT-L/2 (orig) & & 23.33 & $-$ \\
DiT-XL/2 (orig) & & 19.47 & $-$ \\
~+ $7$M steps (orig) & & 12.03 & $-$ \\
\bottomrule
\end{tabular}
}
\vspace{-1em}
\end{table}

%% file: tables/kinetics.tex
\begin{table}[t]
\caption{Results on Kinetics-700~\cite{kinetics700} for DiT trained on top of various autoencoders. See \cref{sec:experiments:main} for details.}
\label{table:k700}
\centering
\resizebox{1.0\linewidth}{!}{
\begin{tabular}{llccc}
\toprule
Stage II & Stage I & \fvd$\downarrow$ & \dinofid$\downarrow$ & \fidtenk$\downarrow$ \\
\midrule
\multirow{3}{*}{DiT-B/2} & \cvae (vanilla) & 650.40 & 650.97 & 28.85 \\
& \cvae + FT & \cellsecond{447.26} & \cellsecond{593.02} & \cellsecond{19.45} \\
& \cvae + FT-\regshortname \ours & \cellbest{252.26} & \cellbest{466.15} & \cellbest{12.19} \\
\midrule
\multirow{3}{*}{DiT-XL/2} & \cvae (vanilla) & \cellsecond{268.26} & 407.23 & \cellsecond{12.02} \\
& \cvae + FT & 270.66 & \cellsecond{402.91} & 12.78 \\
& \cvae + FT-\regshortname \ours & \cellbest{135.15} & \cellbest{245.27} & \cellbest{8.59} \\
\midrule
\multirow{3}{*}{DiT-B/1} & \ltxae (vanilla) & \cellsecond{854.47} & \cellsecond{814.49} & 50.99 \\
& \ltxae + FT & 876.61 & 823.71 & \cellsecond{50.18} \\
& \ltxae + FT-\regshortname \ours & \cellbest{389.56} & \cellbest{642.80} & \cellbest{22.88} \\
\bottomrule
\end{tabular}
}
\vspace{-2em}
\end{table}

%% file: tables/kl-vs-downreg.tex
\begin{table}
\caption{Ablating KL regularization weight $\beta$ for FluxAE fine-tuning in terms of the reconstruction quality and downstream DiT-S/2 and DiT-L/2~\cite{DiT} training. Increasing the KL in general improves the LDM's performance for smaller models, but at the expense of worsened AE reconstruction (and reduced training stability), which can limit scalability (also observed by \cite{SD3}) and results in worse performance of DiT-L/2. Our \regshortname regularization leads to improved LDM performance without hurting the reconstruction and scales well to larger models.}
\label{table:kl-vs-downreg}
\centering
\resizebox{1.0\linewidth}{!}{
\begin{tabular}{lccc}
\toprule
Method & DiT-S/2 \dinofidfivek & DiT-L/2 \dinofidfivek & AE \psnrsmall \\
\midrule
FluxAE (vanilla) & 992.05 & 415.87 & 30.20 \\
~+ KL $\beta = 0$ & 968.26 & 472.08 & 29.97 \\
~+ KL $\beta = 10^{-7}$ & 1018.6 & 425.35 & \cellsecond{30.29} \\
~+ KL $\beta = 10^{-6}$ & 1095.2 & 612.12 & \textcolor{crimsonred}{19.66} \\
~+ KL $\beta = 10^{-5}$ & 940.13 & \cellsecond{403.99} & 29.21 \\
~+ KL $\beta = 10^{-4}$ & 974.67 & 404.61 & 30.22 \\
~+ KL $\beta = 10^{-3}$ & 982.91 & 425.24 & 29.51 \\
~+ KL $\beta = 10^{-2}$ & 1946.5 & 1737.47 & \textcolor{crimsonred}{10.82} \\
~+ KL $\beta = 10^{-1}$ & \cellsecond{929.58} & 472.74 & \textcolor{crimsonred}{23.72} \\
\midrule
~+ FT-SE \ours & \cellbest{924.28} & \cellbest{369.15} & \cellbest{30.37} \\
\bottomrule
\end{tabular}
}
\vspace{-1.5em}
\end{table}

%% file: figures/reg-dct-cut-flux.tex
\begin{figure*}
\centering
\includegraphics[width=\linewidth]{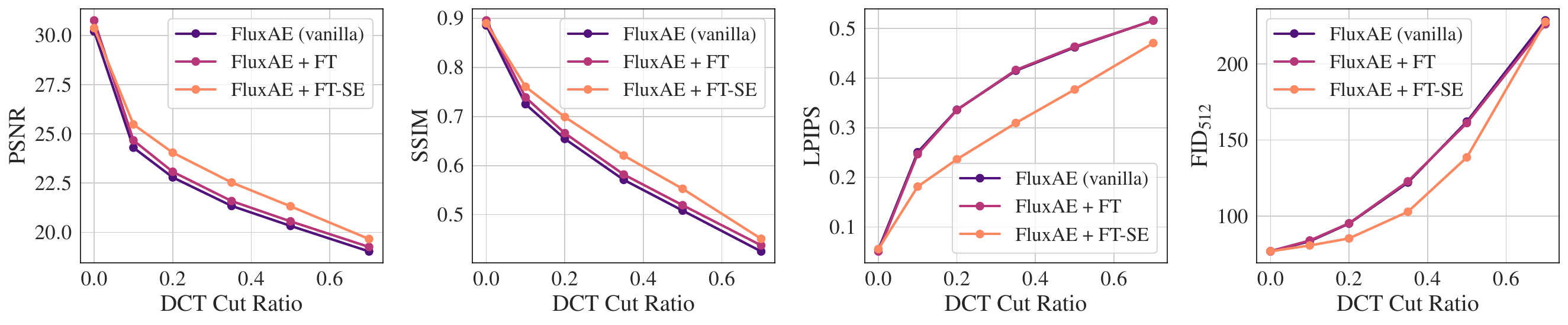}
\vspace{-2em}
\caption{\textbf{Effect of DCT spectrum cutting.} We plot reconstruction metrics on ImageNet for the baseline FluxAE and its finetuned version, with and without \regname. As more high-frequency DCT coefficients are removed from the latents, the AE with regularization consistently achieves the best reconstruction quality.}
\label{fig:reg-dct-cut-flux}
\vspace{-1em}
\end{figure*}

%% file: figures/ablations-downreg.tex
\begin{figure}
\centering
\includegraphics[width=\linewidth]{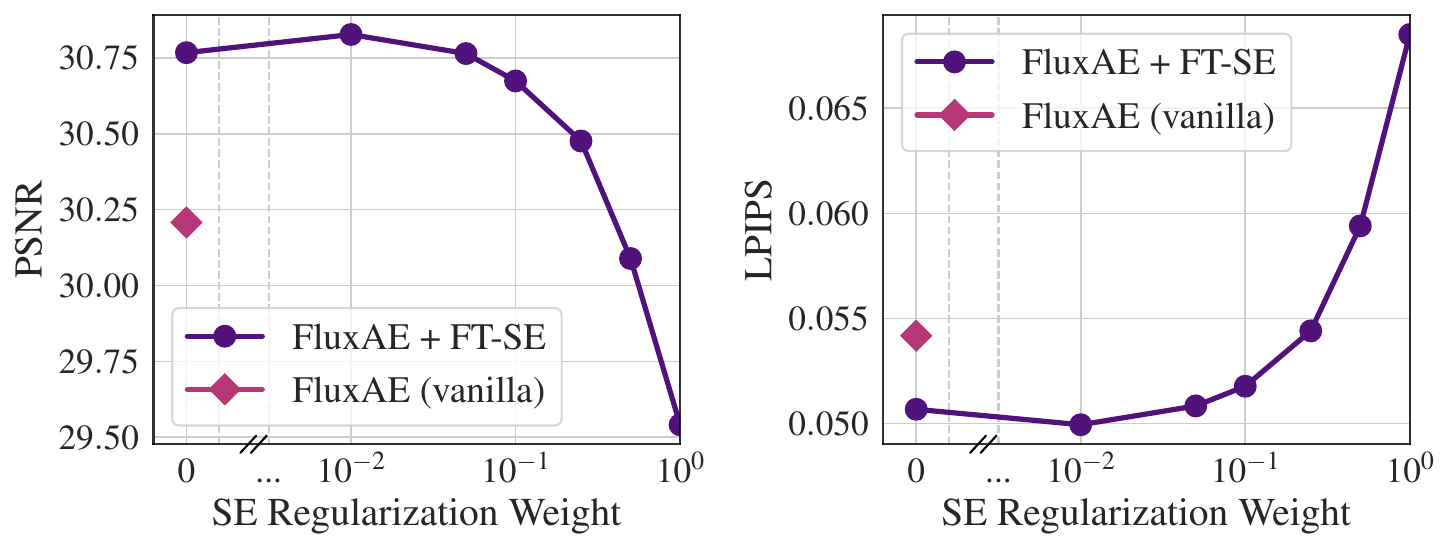}
\vspace{-2em}
\caption{Reconstruction quality with varying \regshortname regularization strength for fine-tuning. We find the value of $0.25$ to be optimal in maintaining reconstruction quality compared to the Base while also improving generation quality as shown in~\cref{table:imagenet,table:k700}.}
\label{fig:ablations-downreg}
\vspace{-1em}
\end{figure}

%% file: tables/ae-rec.tex
\begin{table}
\caption{Influence of spectrum regularization on the reconstruction quality for image and video autoencoders. Image autoencoders are evaluated on 50,000 images from ImageNet-1K, while the video ones are on 50,000 videos from Kinetics-700.}
\label{table:ae-rec}
\centering
\resizebox{1.0\linewidth}{!}{
\begin{tabular}{lccccc}
\toprule
Method & PSNR $\uparrow$ & SSIM $\uparrow$ & LPIPS $\downarrow$ & \fid$\downarrow$  & \fvdfull$\downarrow$ \\
\midrule
FluxAE & 30.243 & 0.883 & \cellbest{0.054} & \cellbest{0.183} & $-$ \\
~+FT-\regshortname \ours & \cellbest{30.474} & \cellbest{0.888} & 0.055 & 0.550 & $-$ \\
\midrule
\cmsaei & 23.230 & 0.638 & 0.181 & \cellbest{1.077} & $-$ \\
~+FT-\regshortname \ours & \cellbest{24.558} & \cellbest{0.677} & \cellbest{0.166} & 1.570 & $-$ \\
\midrule
\cvaefull & 34.961 & 0.947 & 0.073 & 2.992 & 4.614 \\
~+FT-\regshortname \ours & \cellbest{35.399} & \cellbest{0.948} & \cellbest{0.067} & \cellbest{2.986} & \cellbest{3.328} \\
\midrule
\ltxae & \cellbest{30.897} & \cellbest{0.886} & 0.152 & 5.928 & 36.783 \\
~+FT-\regshortname \ours & 30.386 & 0.885 & \cellbest{0.137} & \cellbest{5.303} & \cellbest{34.148} \\
\bottomrule
\end{tabular}
}
\vspace{-1em}
\end{table}

%% file: figures/ts-abl.tex
\begin{figure}[t]
\centering
\includegraphics[width=\linewidth]{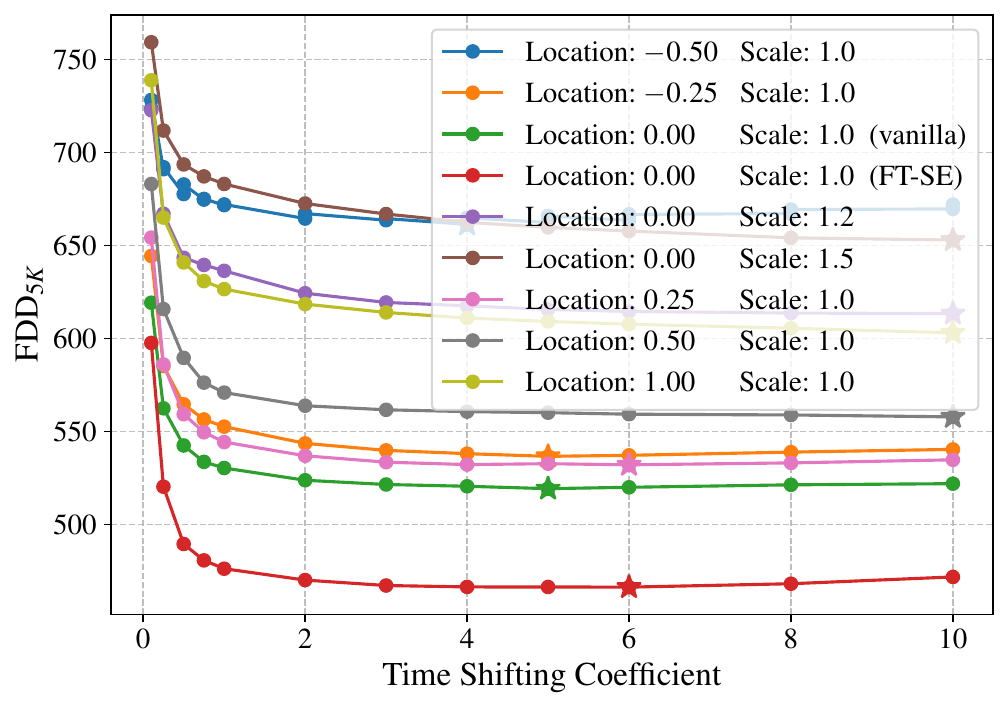}
\vspace{-2em}
\caption{Inference-time time shifting and train-time noise schedule ablation for DiT-B/2 models trained on top of vanilla FluxAE, and also on top of our regularized autoencoder. Our regularized FluxAE-FT+\regshortname achieves the best results across \emph{all} the setups, confirming that our improved quality is not due to improved diffusability, rather than implicit time shifting. The optimal time shifting coefficient is marked with a star.}
\label{fig:time-shifting-abl}
\vspace{-1.5em}
\end{figure}

%% file: sections/5-conclusion.tex
\section{Conclusion}
\label{sec:conclusion}


We have shown that modern Latent Diffusion Models (LDMs) rely just as critically on their autoencoders as on the more frequently investigated diffusion architectures.
While prior work has largely focused on improving reconstruction quality and compression rates for the autoencoders, our study focuses on \diffusability, revealing how latent spectra with excessive high-frequency components can hamper the downstream diffusion process.
Through a systematic analysis of several autoencoders, we uncovered stark discrepancies between latent and RGB spectral properties and demonstrated that they lead to worse LDM synthesis quality.
Building on this insight, we developed a regularization strategy that aligns the latent and RGB spaces across different frequencies.
Our approach maintains reconstruction fidelity and improves diffusion training by suppressing spurious high-frequency details in the latent code.
Potential future directions include exploring more advanced frequency-based regularizations, adaptive compression methods and scale equivariance regularization in the temporal axis for video autoencoders to further optimize the trade-off between reconstruction quality, compression rate, and diffusability.


%% file: sections/6-impact.tex
\section*{Impact Statement}

This work focuses on improving the representations in autoencoders that serve as a backbone for latent diffusion training, ultimately enhancing generative performance.
Our improvements can facilitate beneficial applications such as boosting creativity, supporting educational content creation, and reducing computational overhead in generative workflows.
Beyond these considerations, we do not identify additional ethical or societal implications beyond those already known to accompany large-scale generative modeling.

%% file: supp/1-limitations.tex
\section{Limitations.}

We identify the following limitations of our work and the proposed regularization:
\begin{enumerate}
    \item While we did our best to verify that our framework works in the most general setup possible, testing 4 different autoencoders across 2 different domains (image and videos), our study would be more complete when verified across other diffusion parametrizations~\cite{EDM, DDPM, VDM++} or architectures~\cite{EDMv2}.
    \item We observed that our regularization still affects the reconstruction slightly: for example, \cref{table:ae-rec} shows that FluxAE \fid increased from 0.183 to 0.55 (though for some AEs, like \cvaefull, it improves). We are convinced that this \fid increase could be mitigated by training with adversarial losses, which we omitted in this work for simplicity.
    \item There is a mild sensitivity to hyperparameters: for example, we found that varying the SHF regularization weight might improve the results (see \cref{table:regweight-abl}), or adding a small KL regularization (which we disabled in the end for our regularization for simplicity).
    \item None of the explored autoencoders released their training pipelines, and it is non-trivial to fine-tune them even without any extra regularization. For example, we observed that any fine-tuning of \dcae~\cite{DC-AE} was leading to divergent reconstructions in our training pipeline (we explored dozens of different hyperparameter setups). 
\end{enumerate}

We leave the exploration of these limitations for future work.

%% file: supp/2-details.tex
\section{Implementation Details}
\label{ap:details}

\inlinesection{DiT model details}.
To strengthen the baseline DiT performance, we integrated into it the latest advancements from the diffusion model literature.
Namely, we used self conditioning~\cite{RIN} and RoPE~\cite{RoPE} positional embeddings.
Besides, we switched to the rectified flow diffusion parametrization~\cite{NormFlowsWithStochInterp, RecFlow, LSGM}, which was recently shown to have better scalability with a fewer amount of inference steps~\cite{SD3}.

Following \cite{RIN,FIT,WALT,SnapVideo}, we employ \emph{self-conditioning} for our DiT training and inference.
During training with a 90\% probability, we run an auxiliary forward pass with the DiT model, take its activations from the last block (i.e., right before the ``unpatchify'' projection), project them with a linear layer and add as residuals to the input tokens after patchification of the main training forward pass.
For that auxiliary forward pass, following RIN~\cite{RIN}, we use the same noise level $\sigma$ and ``no-grad'' context (i.e., we do not backpropagate through the auxiliary forward pass).

\inlinesection{DiT training details}.
All the DiT models are trained for 400,000 steps with 10,000 warmup steps of the learning rate from 0 to 0.0003 and then its gradual decay towards 0.00001.
We used weight decay of 0.01 and AdamW~\cite{AdamW} optimizer with beta coefficients of 0.9 and 0.99.
We used posterior sampling from the encoder distribution for VAE-based autoencoders.
In contrast to the original work, we found it helpful to do learning rate decay to 0.00001 using the cosine learning rate schedule.
We used the same model sizes for DiT-S (small), DiT-B (base), DiT-L (large) and DiT-XL (extra large), as the original work~\cite{DiT}:
\begin{itemize}
    \item DiT-S: hidden dimensionality of 384, 12 transformer blocks, and 6 attention heads in the multi-head attention.
    \item DiT-B: hidden dimensionality of 768, 12 transformer blocks, and 12 attention heads in the multi-head attention.
    \item DiT-L: hidden dimensionality of 1024, 24 transformer blocks, and 16 attention heads in the multi-head attention.
    \item DiT-XL: hidden dimensionality of 1152, 28 transformer blocks, and 16 attention heads in the multi-head attention.
\end{itemize}
We used gradient clipping with the norm of 16 for all the DiT models.
Our models were trained in the FSDP~\cite{FSDP} framework with the full sharding strategy on a single node of $8 \times$ NVidia A100 80GB GPUs or $8 \times$ NVidia H100 80GB GPUs (depending on their availability in our computational cluster).

For \cvae, since it is considerably slower than other autoencoders, we trained LDMs on pre-extracted latents.
For this, we pre-extracted them on random 17-frames clips.
In essence, this reduces the total dataset size, but since we do the same procedure for the entire CogVideoX-AE family, the models are comparable between each other.

\inlinesection{Autoencoders training details}.
Since none of the autoencoders had their training pipelines released, we had to develop the training recipes for each of the autoencoder baselines individually which would not be detrimental to neither their reconstruction capability nor downstream diffusion performance.
To do this, we ablated multiple hyperparameters (the most important ones being learning rate and KL regularization strength) to arrive to a proper setup.
We chose the KL weight in such a way that the KL penalty maintains approximately the same magnitude as the pre-trained checkpoint.

Each autoencoder is trained with AdamW~\cite{AdamW} optimizer, with betas of 0.9 and 0.99, and weight decay of 0.01.
The learning rate was grid-searched individually for each autoencoder and is provided in \cref{tab:hyperparameters}.
In all the cases, we used mixed precision training with BFloat16.

During training, we maintained an exponential moving average of the weights~\cite{EDMv2}, initialized from the same parameters as the starting model, and having a half life of 5,000 steps.

We emphasize that, when applying our regularization strategy on top of an autoencoder baseilne, we do not alter other hyperparameters (like learning rate), except for KL regularization which we disable for \regshortname-regularized models (even though we found it helpful in some of our explorations).

For each autoencoder, we freeze the last output layers of the decoder.
The motivation is the following: they were fine-tuned with the adversarial loss, which we want to exclude from the equation without hurting the ability of an autoencoder to model textural details which \fid would be sensitive to~\citep{LDM} and which do not influence the latent space properties.
Namely, we freeze the last normalization and output convolution layers.
In each case, the amount of frozen parameters constitute a negligible amount of total parameters.

Other hyperparameters for autoencoders training are provided in \cref{tab:hyperparameters}.

\input{tables/hyperparameters}

\inlinesection{Time shifting for rectified flow}.
In \cref{sec:experiments:abl}, we discussed the influence of time shifting on the DiT model.
Here, we provided the details of its implementation.
We adapt the time shifting schedule from Lumina-T2X~\cite{Lumina-T2X} for rectified flow parametrization by rescaling the time steps at inference time using the following formula:
\begin{equation}
t' = 1 - \frac{1 - t}{s + (1 - s) \cdot (1 - t)},
\end{equation}
where $t$ is the time step, going from 1 (full noise) to 0 (clean image), and $s$ is the shifting coefficient.
One can see that for $s = 1$, $t' = t$, which is the default unshifted behaviour.

%% file: tables/hyperparameters.tex
\begin{table*}[h]
\caption{Hyperparameters for the autoencoders explored in the current work. We had to tweak the hyperparameters for various autoencoders to prevent the divergence of the baseline training.}
\label{tab:hyperparameters}
\centering
\begin{tabular}{lcccc}
\toprule
Hyperparameter & FluxAE & \cmsaei & \cvae & \ltxae \\
\midrule
Domain & image & image & video & video \\
Compression rate & $8 \times 8$ & $16 \times 16$ & $4 \times 8 \times 8$ & $8 \times 32 \times 32$ \\
Latent channels & 16 & 16 & 16 & 32 \\
Number of fune-tuning steps & 10,000 & 10,000 & 20,000 & 20,000 \\
Image batch size & 32 & 32 & 64 & 64 \\
Video batch size & 0 & 0 & 32 & 32 \\
Default KL $\beta$ weight & 0.001 & 0.0 & 0.001 & 0.0001 \\
Learning rate & 0.00001 & 0.0001 & 0.0003 & 0.00005 \\
Number of parameters & 83.8M & 44M & 211.5M & 419M \\
Training resolution & $256 \times 256$ & $256 \times 256$ & $17 \times 256 \times 256$ & $17 \times 256 \times 256$ \\
MSE loss weight & 1 & 1 & 1 & 5 \\
LPIPS loss weight & 1.0 & 1.0 & 1.0 & 1.0 \\
Gradient clipping norm & 50 & 50 & 1 & 50 \\
Num upsampling blocks frozen & 1 & 3 & 0 & 0 \\
Is output convolution frozen? & Yes & Yes & Yes & Yes \\
\bottomrule
\end{tabular}
\end{table*}

%% file: supp/3-extra-exps.tex
\section{Additional Experiments}
\label{ap:freq-reg}

In \cref{sec:method}, we outlined the base \regname strategy to regularize the spectrum of an autoencoder which has a strong advantage of being very easy to implement by a practitioner.
However, it could be beneficial to possess more advanced tools for a finer-grained control over the latent space spectral properties.
This section outlines them and provides the corresponding ablation.

\subsection{Explicitly Chopping off High-Frequency Components}

Rather than applying downsampling to produce latents and RGB targets for regualrization, it is possible to replace some ratio of high-frequency components with zeros. To do so, DCT is applied to the latents and RGB targets where a chosen set of frequency components are masked out. The modified components are then translated back to the spatial domain by inverse DCT to form the training latents and reconstruction targets.

\begin{equation}\label{eq:hard-hf-penalty}
\loss_\text{CHF}(x) = d(x, \Dec(z)) + d( {D^{-1}}(D({x}) * \mathbf{M}), \Dec({D^{-1}}(D({z}) * \mathbf{M}) ) + \loss_\text{reg},
\end{equation}
where $D$ and $D^{-1}$ represent DCT and its inverse, respectively. $\mathbf{M}$ is a $B \times B$ binary mask indicating which frequencies to zero out defined as follows:
\begin{equation}\label{eq:zigzag-mask}
\mathbf{M}(u,v) =
\begin{cases}
    1, & \text{if } \zigzag (u,v) < B^2 - N,\\
    0, & \text{otherwise}.
\end{cases}
\end{equation}
$N$ controls the frequency cutoff. We provide the ablation for this strategy in \cref{table:cuthf-ablation}.

\begin{table}[h]
\caption{Ablations for explicit high-frequency chop off for DiT-S/2 trained for 200,000 iterations on top of Flux AE with such a regularization. While it can achieve better results for some of the baselines than naive downsampling, we opt out for the latter strategy due to its simplicity. For the non-zigzag order ablation, we cut across each $x$ and $y$ axes independently}
\label{table:cuthf-ablation}
\centering
\begin{tabular}{llcc}
\toprule
Stage II & Stage I  & \dinofidfivek \\
\midrule
DiT-S/2 & FluxAE + chop off 90\% (non-zigzag order) & 912.4 \\
DiT-S/2 & FluxAE + chop off 70\% (non-zigzag order) & 915.6 \\
DiT-S/2 & FluxAE + chop off 30\% (non-zigzag order) & 929.7 \\
DiT-S/2 & FluxAE + chop off 10\% (non-zigzag order) & 916.5 \\
DiT-S/2 & FluxAE + chop off 90\% (zigzag order) & 935.5 \\
DiT-S/2 & FluxAE + chop off 70\% (zigzag order) & 932.8 \\
DiT-S/2 & FluxAE + chop off 30\% (zigzag order) & 962.9 \\
DiT-S/2 & FluxAE + chop off 10\% (zigzag order) & 930.1 \\
\midrule
DiT-S/2 & FluxAE (vanilla) & 992.0 \\
DiT-S/2 & FluxAE with optimal (out of 8) KL $\beta$ & 929.6 \\
\bottomrule
\end{tabular}
\end{table}

In \cref{fig:progressive-dct-cut}, we provided the visualizations for a FluxAE resiliense with and without such chopping high-frequency regularization for $50\%$ HF dropout rate.
In \cref{fig:progressive-dct-cut-downreg}, we provide an equivalent visualization for \regshortname-fine-tuned FluxAE: while it is less resilient to frequency dropout than \regchfshortname, but is still noticeably better than the vanilla model.

\input{figures/qualitative-dct-cut-downreg}

\subsection{Soft Penalty for High-Frequency Components}

Instead of directly removing some of the components, which might become a too strict regularization signal, one can consider penalizing the amplitudes of high-frequency components in a soft manner.
Concretely, given a $B \times B$ block, we construct the following weight penalty matrix:

\begin{equation}\label{eq:soft-hf-penalty-matrix}
\mathcal{W}_{uv} = (u + v)^p / B^p.
\end{equation}

Next, the soft regularization loss itself is computed as:
\begin{equation}\label{eq:soft-hf-penalty}
\loss_\text{softreg} = \sum_{u,v} D_{uv}(z) \cdot \mathcal{W}_{uv}.
\end{equation}

During training, when enabled, we add it to the main loss with the weigh $\gamma$.
We found it beneficial in some of our experiments when it is added with a small coefficient (e.g., 0.01).
While it is possible to achieve higher results with more fine-grained regularization, we opt to use the simpler version since we believe it would be easier to employ by the community.

To ablate its importance, we trained DiT-B/2 model on top of FluxAE models, fine-tuned with a different strength $\gamma$.
The results are presented in \cref{table:softregweight-abl}.

\begin{table}[h!]
\caption{Ablating the regularization strength $\gamma$ of the soft frequency regularization $\loss_\text{softreg}$.}
\label{table:softregweight-abl}
\centering
\begin{tabular}{llcc}
\toprule
Stage II & Stage I  & \fid$_{5k}$ & \dinofid$_{5k}$ \\
\midrule
\multirow{6}{*}{DiT-B/2}
& FluxAE + FT-\regshortname $\gamma = 0.001$ & 26.43 & 497.14 \\
& FluxAE + FT-\regshortname $\gamma = 0.025$ & 25.46 & 477.61 \\
& FluxAE + FT-\regshortname $\gamma = 0.01$ & 26.72 & 487.06 \\
& FluxAE + FT-\regshortname $\gamma = 0.05$ & \cellbest{24.28} & \cellbest{458.11} \\
& FluxAE + FT-\regshortname $\gamma = 0.1$ & \cellsecond{25.84} & \cellsecond{461.97} \\
\bottomrule
\end{tabular}
\end{table}

\begin{figure}[h!]
\centering
\includegraphics[width=0.4\linewidth]{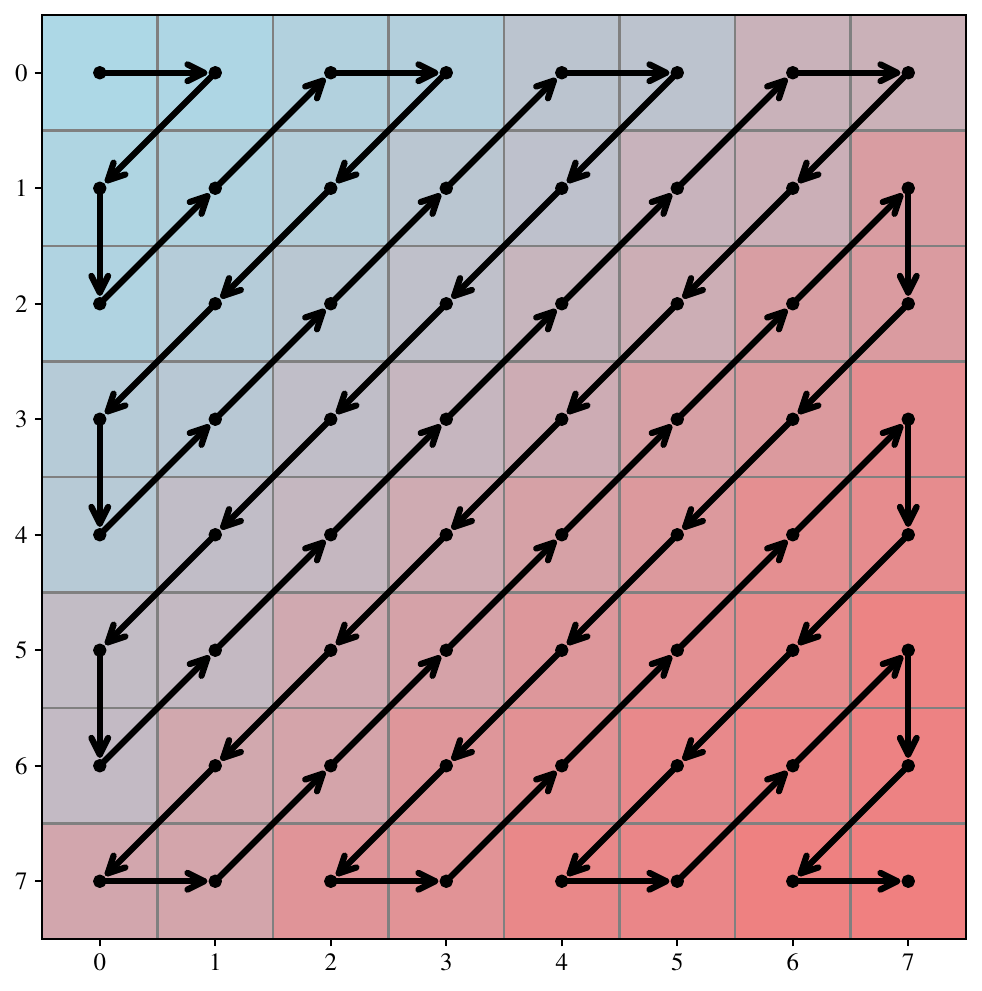}
\caption{Illustration of the zigzag indexing order of DCT.}
\label{fig:zigzag}
\end{figure}

\subsection{ImageNet $512^2$ experiments}

We trained our DiT-L/2 for class-conditional $512^2$ ImageNet-1K generation for 400K steps for FluxAE~\cite{Flux}, the results are presented in \cref{table:imagenet-512}.

\input{tables/imagenet-512}

\subsection{Ablating regularization strength $\alpha$}

To ablate the importance of the regularization strength $\alpha$, we train FluxAE for 10,000 steps with a varying strength.
The results are presented in \cref{table:regweight-abl}.

\begin{table}[h!]
\caption{Ablating the regularization strength $\alpha$ of our proposed \regname regularization.}
\label{table:regweight-abl}
\centering
\begin{tabular}{llcc}
\toprule
Stage II & Stage I  & \fid$_{5k}$ & \dinofid$_{5k}$ \\
\midrule
\multirow{6}{*}{DiT-B/2}
& FluxAE + FT-\regshortname $\alpha = 0.01$ & 33.99 & 641.95 \\
& FluxAE + FT-\regshortname $\alpha = 0.05$ & 33.86 & 645.94 \\
& FluxAE + FT-\regshortname $\alpha = 0.1$ & \cellsecond{28.62} & \cellsecond{586.91} \\
& FluxAE + FT-\regshortname $\alpha = 0.25$ & \cellbest{26.84} & \cellbest{558.36} \\
& FluxAE + FT-\regshortname $\alpha = 0.5$ & 29.63 & 569.92 \\
& FluxAE + FT-\regshortname $\alpha = 1$ & 33.22 & 612.45 \\
\bottomrule
\end{tabular}
\end{table}

\subsection{From scratch training}
\label{ap:sec:from-scratch}
To conduct whether our regularization provides benefits for from scratch training and when the same data is being used for both AE and LDM, we conducted additional experiments for \fluxae and \cvaefull. We trained FluxAE from scratch on ImageNet $256^2$ with/without our regularization and then trained DiT-B/2 on top of the latent spaces of these two autoencoders.
Then, we did the same for \cvaefull for Kinetics-700 $17 \times 256^2$.
The results of AE reconstruction performance (after 300K training steps for \fluxae and 60K training steps for \cvaefull) and LDM generation performance after 400K training steps are presented in \cref{tab:from-scratch}.
One can observe that our regularization improves the performance in both cases.

\begin{table}[h!]
\caption{From scratch training results for \fluxae~\cite{Flux} and \cvaefull~\cite{CogVideoX} See \cref{ap:sec:from-scratch} for details.}
\label{tab:from-scratch}
\centering
\begin{tabular}{lccccc}
\toprule
Method & PSNR $\uparrow$ & LPIPS $\downarrow$ & g\dinofidfivek$\downarrow$ & g\fidfivek$\downarrow$ & g\fvdfivek$\downarrow$ \\
\midrule
\fluxae & \cellbest{29.39} & \cellbest{0.059} & 502.44 & 25.43 & $-$ \\
~+FT-\regshortname \ours & 28.73 & 0.065 & \cellbest{472.26} & \cellbest{22.32} & $-$ \\
\midrule
\cvaefull & \cellbest{37.08} & \cellbest{0.049} & 614.95 & 19.87 & 336.33 \\
~+FT-\regshortname \ours & 36.61 & 0.053 & \cellbest{515.68} & \cellbest{17.40} & \cellbest{237.28} \\
\bottomrule
\end{tabular}
\end{table}

\subsection{Compute cost analysis}
One might argue that the improved performance of our regularization comes from the extra computation which the autoencoder is using during training.
In this section, we show that it is not the case by providing an analysis of its computational cost.

We measured the FLOPs of FluxAE (for the batch size of 1 and resolution of 2562) using the popular \href{https://github.com/facebookresearch/fvcore}{fvcore} library.
The entire encoder-decoder pass has 447 GFLOPS, and is split between the encoder/decoder as 136 vs 311 GFLOPS.
Our regularization reuses the encoder pass and only runs the decoder with $\times 2$ or $\times 4$ reduced resolution (the scale sampled randomly during training).
This results in 77.6 or 19.4 extra GFLOPs of the decoder, which is almost exactly 1/4 or 1/16 of the decoder compute or +17\% or 4.5\% of the total forward pass.
Since we sample $\times 2$ or $\times 4$ downsampling factor equally randomly, this results in ~10.75\% of total FLOPs overhead for our regularization.

To strengthen our point even further, we ran an experiment where the baseline FluxAE was fine-tuned strictly for 2 times longer (for $20$K instead of $10$K iterations).
The resulted DiT-B/2 model achieved \fidfivek and \dinofidfivek of only 33.99 and 642.7 versus the scores of 25.87 and 551.27 respectively for our regularized FluxAE+FT-SE model, fine-tuned for only 10K steps.

% ...

%% file: figures/qualitative-dct-cut-downreg.tex
\begin{figure}[t]
\centering
\begin{minipage}{0.7\linewidth}
\setlength{\unitlength}{\linewidth}
\centering
\vspace{-2.21cm}
\begin{picture}(1,1)
\put(0.05, 0){\includegraphics[width=0.95\linewidth]{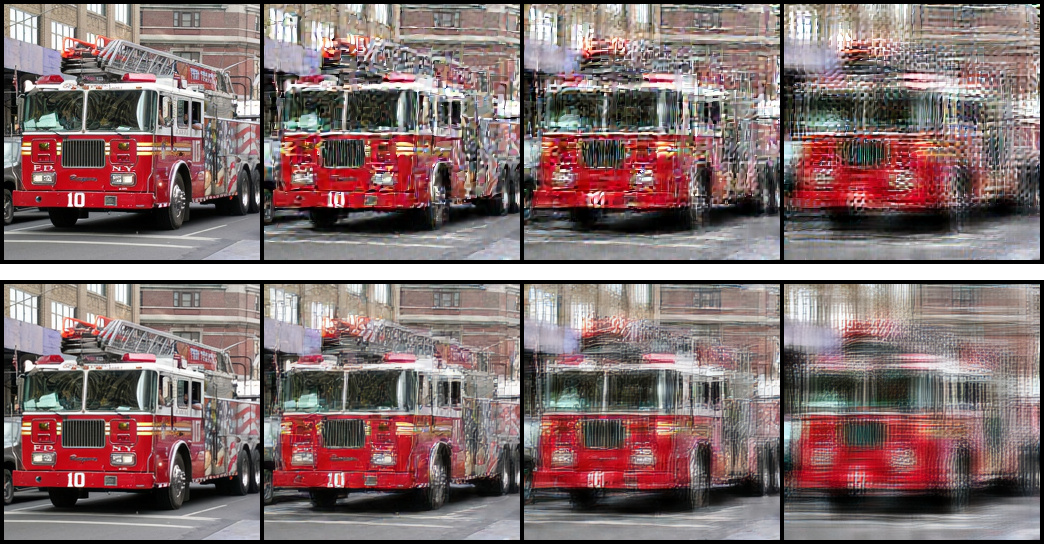}}

\put(0.175, -0.01){\makebox(0,0)[t]{\small 0\%}}
\put(0.415, -0.01){\makebox(0,0)[t]{\small 25\%}}
\put(0.65, -0.01){\makebox(0,0)[t]{\small 50\%}}
\put(0.89, -0.01){\makebox(0,0)[t]{\small 75\%}}

\put(0.035, 0.365){\makebox(0,0)[r]{\rotatebox{90}{\small FluxAE}}}
\put(0.035, 0.12){\makebox(0,0)[r]{\rotatebox{90}{\small FluxAE + \regshortname}}}
\end{picture}
\end{minipage}
\caption{
RGB and FluxAE reconstruction with/without scale equivariance regularization for different percentages of chopped-off high frequency components.
}
\label{fig:progressive-dct-cut-downreg}
\end{figure}

%% file: tables/imagenet-512.tex
\begin{table}[ht]
\caption{Class-conditional generation results on ImageNet-1K $512^2$ without guidance. The original DiT paper reports the results after 3M training steps, while we use 400K steps for our models.}
\label{table:imagenet-512}
\centering
\begin{tabular}{llcc}
\toprule
Stage II & Stage I  & \fid & \dinofid \\
\midrule
\multirow{3}{*}{DiT-L/2} & \fluxae (vanilla) & \cellsecond{13.13} & \cellsecond{249.4} \\
& \fluxae + FT & 13.69 & 267.7 \\
& \fluxae + FT-\regshortname \ours & \cellbest{11.63} & \cellbest{203.5} \\
\midrule
DiT-XL/2 (orig) + 3M steps & SD-VAE-ft-MSE & 12.03 & $-$ \\
\bottomrule
\end{tabular}
\end{table}

%% file: supp/4-visualizations.tex
\section{Additional visualizations}
\label{ap:visualizations}

\subsection{Spectra visualizations}
\label{ap:visualizations:spectra}

\begin{figure}[h!]
\centering
\includegraphics[width=0.5\linewidth]{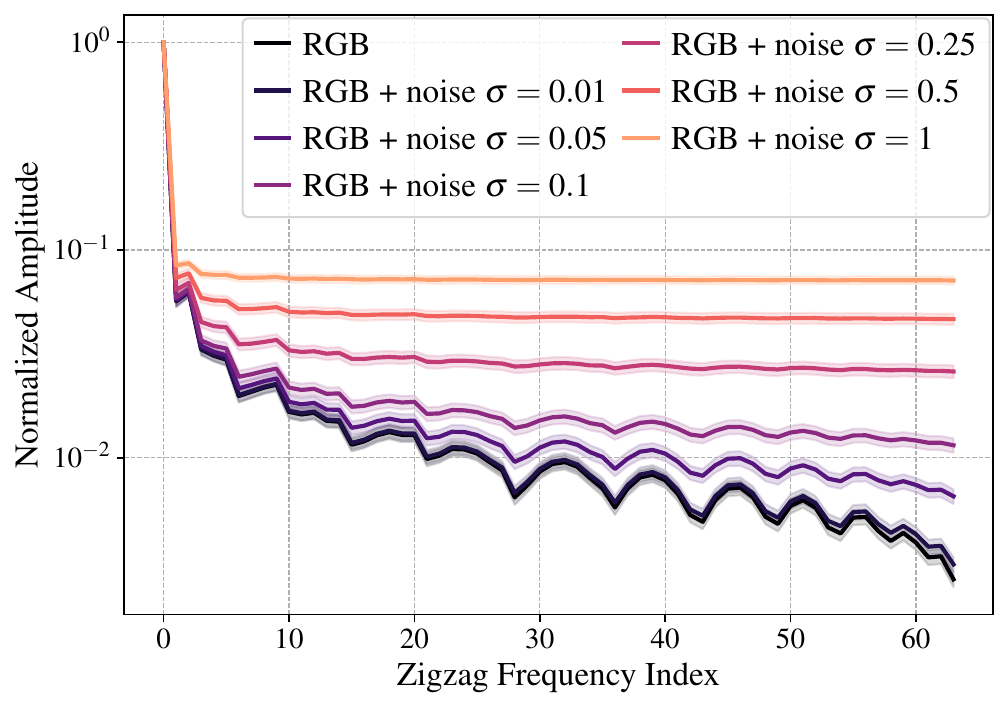}
\caption{Spectrum of RGB under different noising levels: noise inflates the high-frequency component (when normalized). This results into an undesirable side-effect of KL regularization.}
\label{fig:rgb-noise}
\end{figure}

\begin{figure}[ht!]
\centering

\begin{subfigure}[t]{0.45\linewidth}
  \centering
  \includegraphics[width=\linewidth]{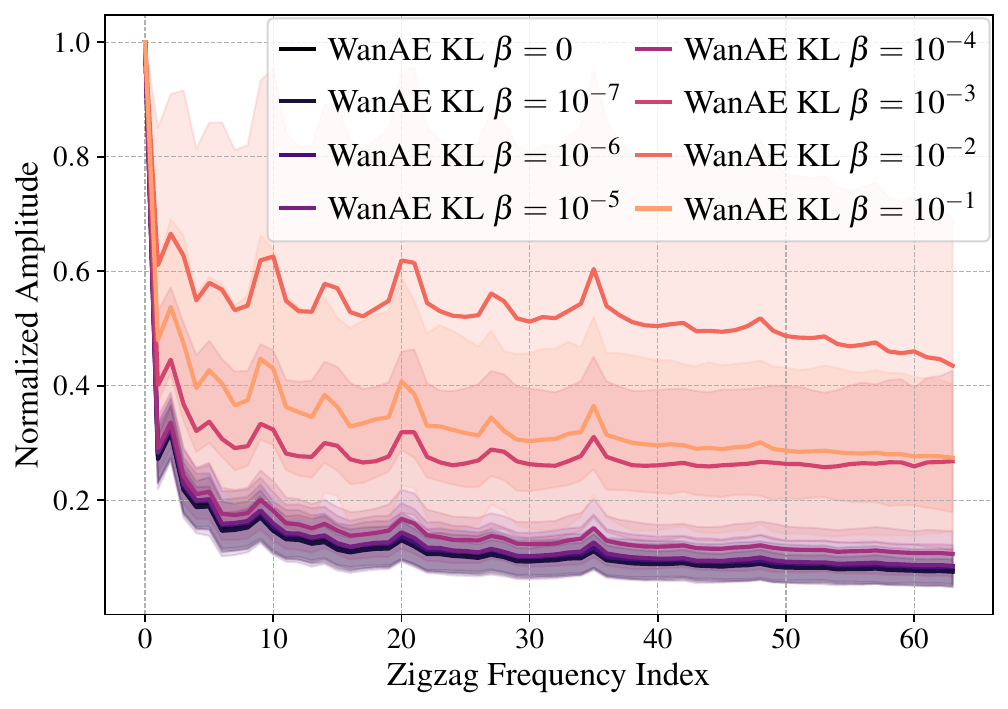}
  \caption{WanAE, KL regularization.}
  \label{fig:wanae-kl}
\end{subfigure}
\hfill
\begin{subfigure}[t]{0.45\linewidth}
  \centering
  \includegraphics[width=\linewidth]{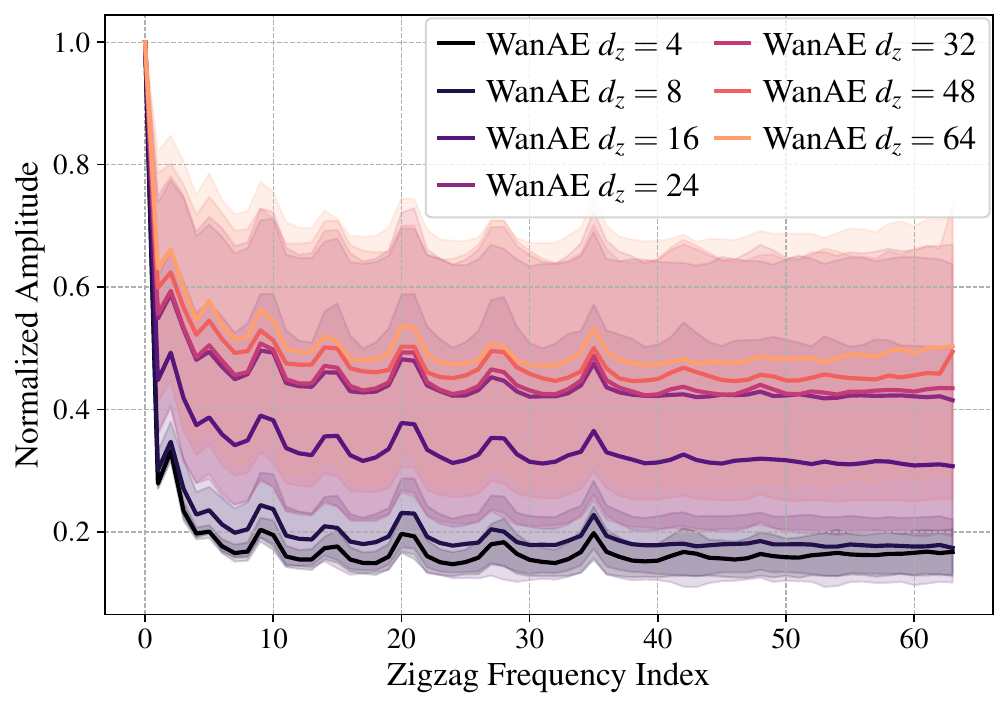}
  \caption{WanAE, channel dimensions.}
  \label{fig:wanae-ch}
\end{subfigure}

\vspace{1em}

\begin{subfigure}[t]{0.45\linewidth}
  \centering
  \includegraphics[width=\linewidth]{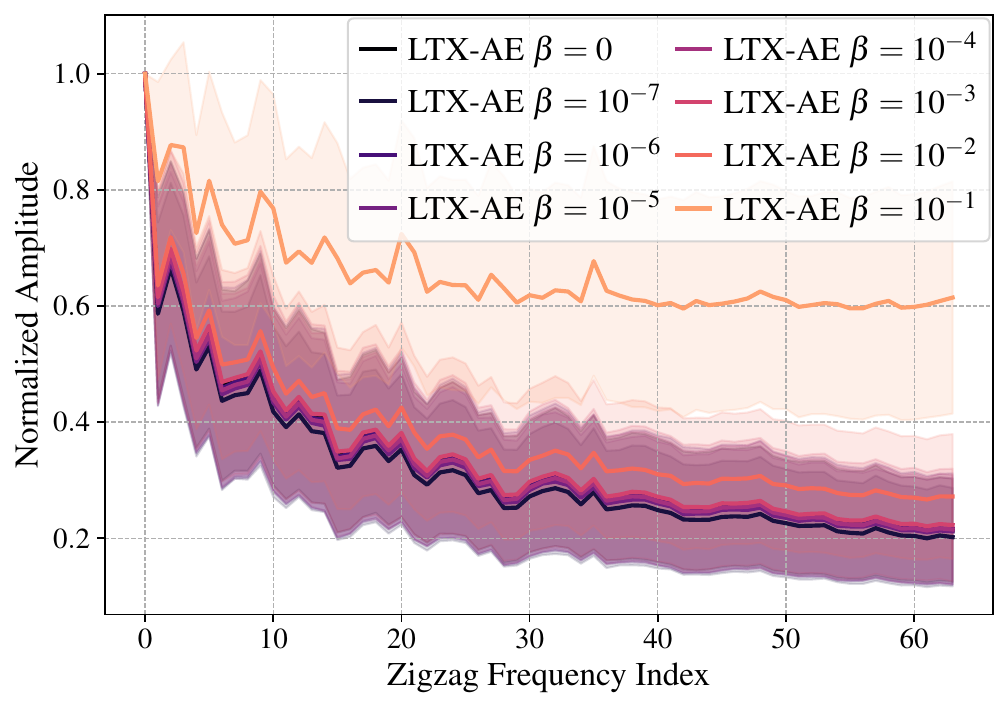}
  \caption{LTX-AE, KL regularization.}
  \label{fig:ltxae-kl}
\end{subfigure}
\hfill
\begin{subfigure}[t]{0.45\linewidth}
  \centering
  \includegraphics[width=\linewidth]{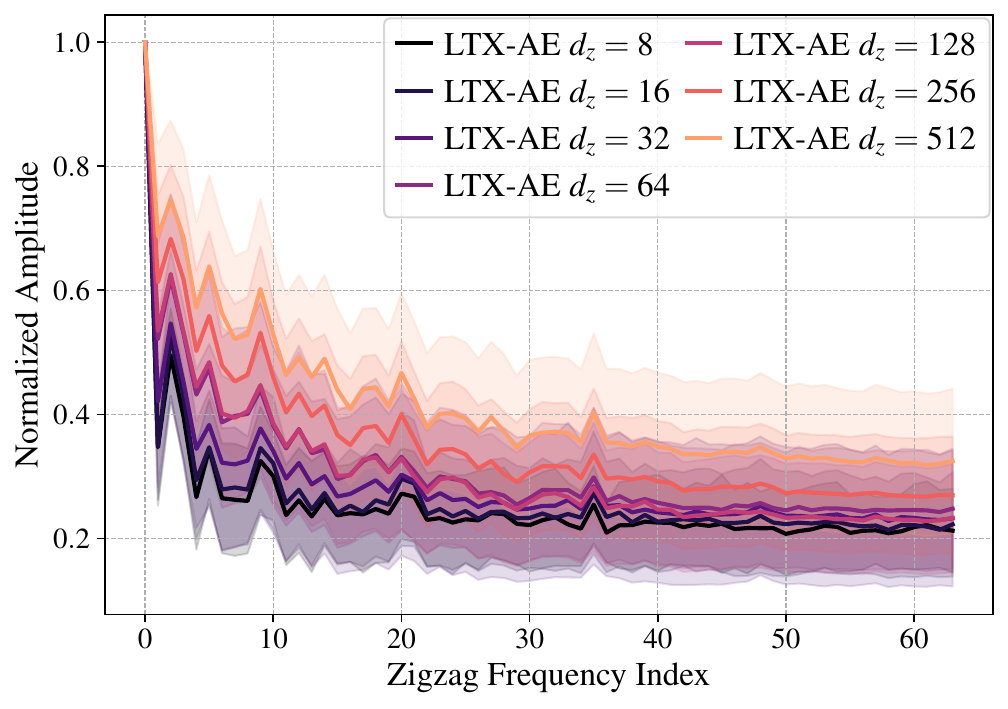}
  \caption{LTX-AE, channel dimensions.}
  \label{fig:ltxae-ch}
\end{subfigure}

\caption{Latent spectra of WanAE~\cite{Wan} and LTX-AE~\cite{LTX-video} trained from scratch on Kinetics-700 $17 \times 256^2$ for 100K steps with varying KL regularization strength $\beta$ or channel size.}
\label{fig:spectra-grid}
\end{figure}


\clearpage
\subsection{Samples visualizations}
\label{ap:visualizations:samples}

\begin{figure*}[h!]
\centering
\includegraphics[width=\linewidth]{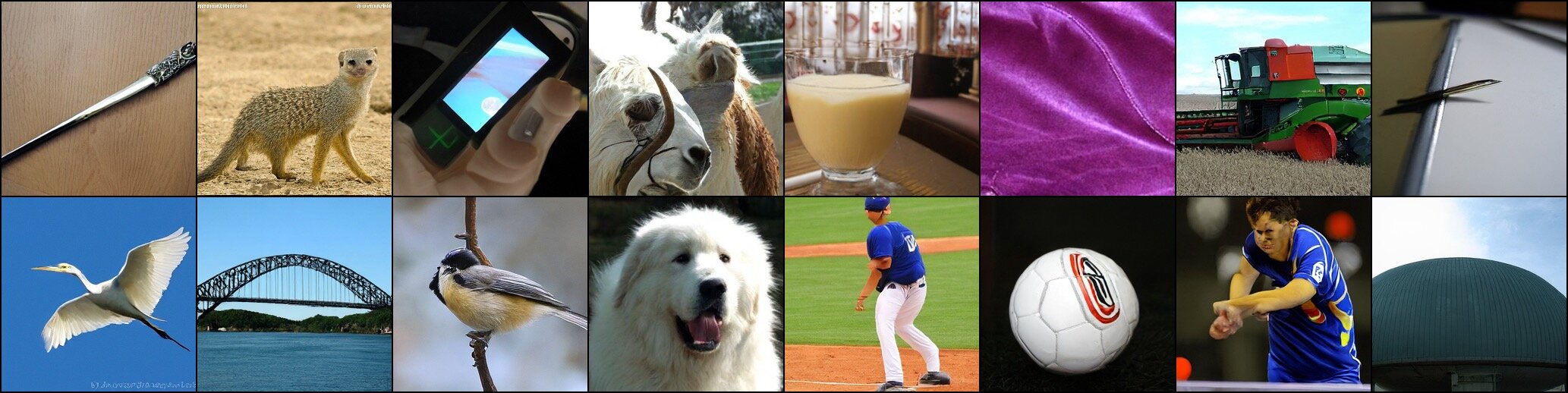}
\includegraphics[width=\linewidth]{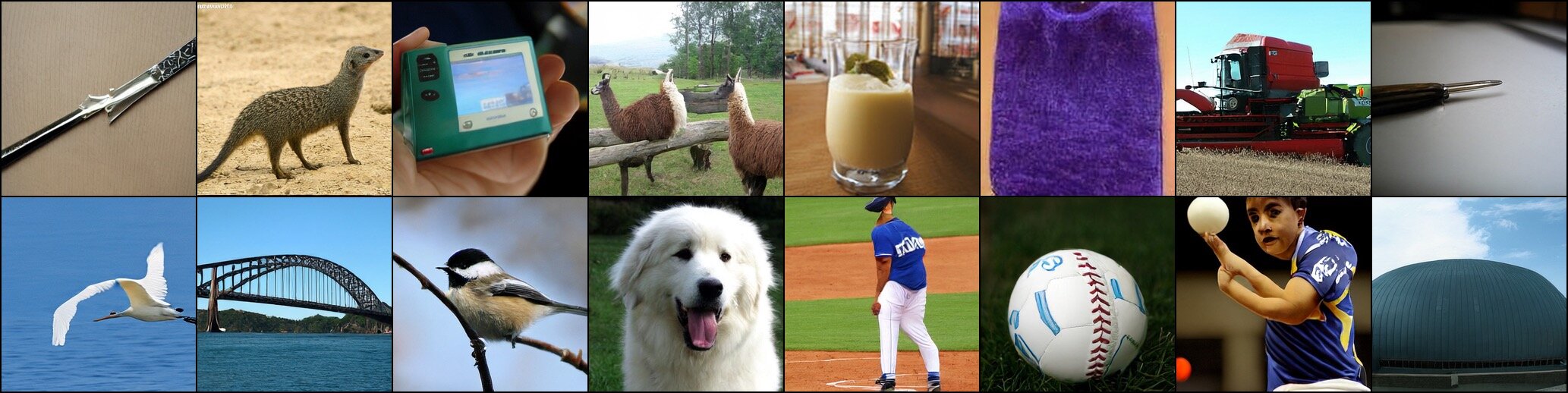}
\includegraphics[width=\linewidth]{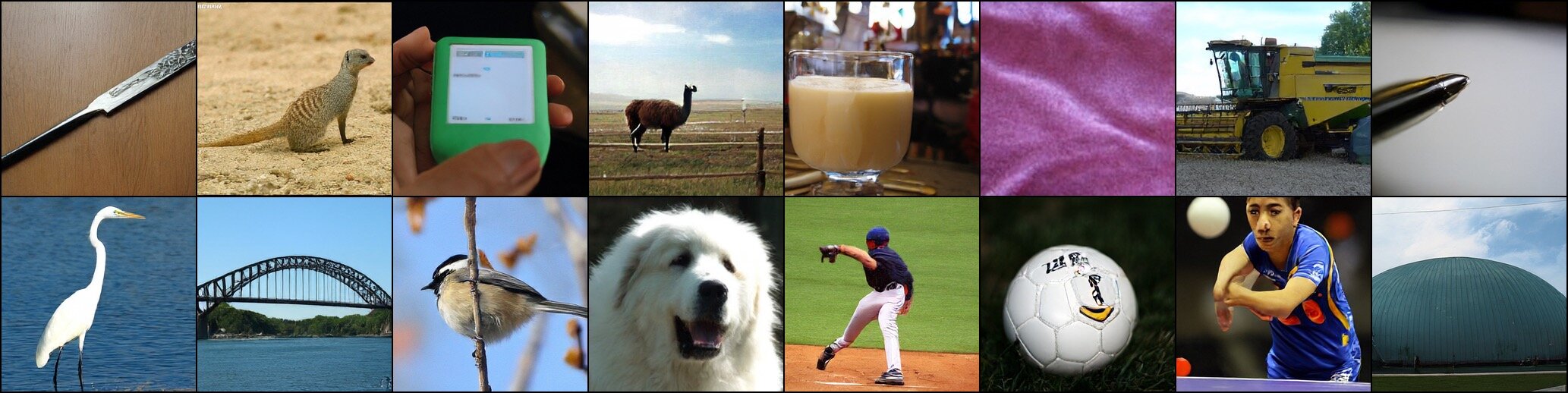}
\caption{Uncurated samples from DiT-XL/2 for FluxAE (top), FluxAE + FT (middle) and FluxAE + \regshortname (bottom) on class-conditional ImageNet $256 \times 256$ for random classes. During inference, we used 256 steps with the guidance scale of 3.0.}
\label{fig:ap:extra-samples-ditxl-flux}
\end{figure*}

\begin{figure*}
\centering
\includegraphics[width=\linewidth]{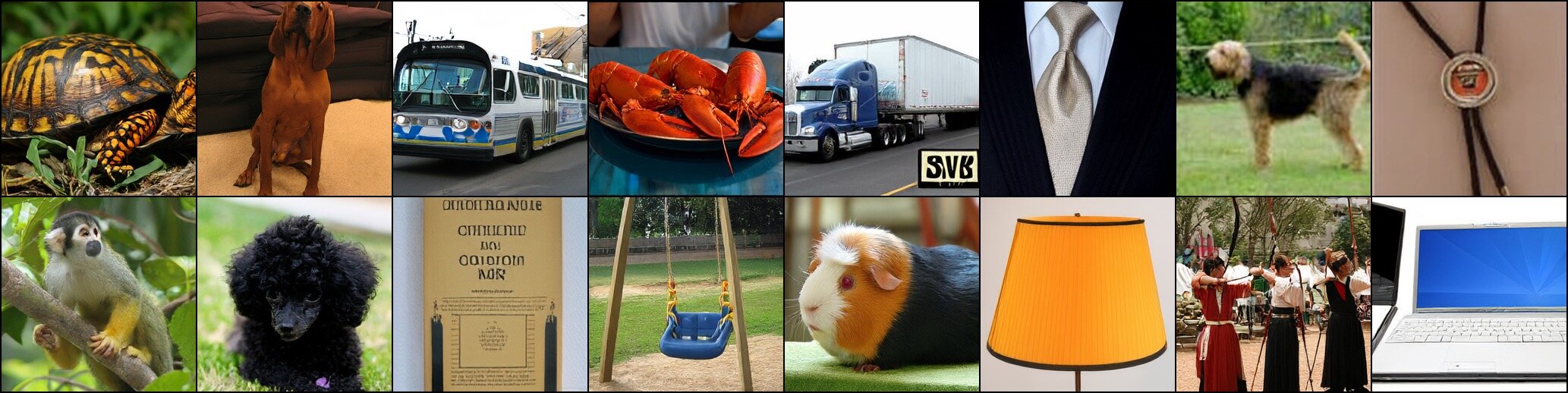}
\includegraphics[width=\linewidth]{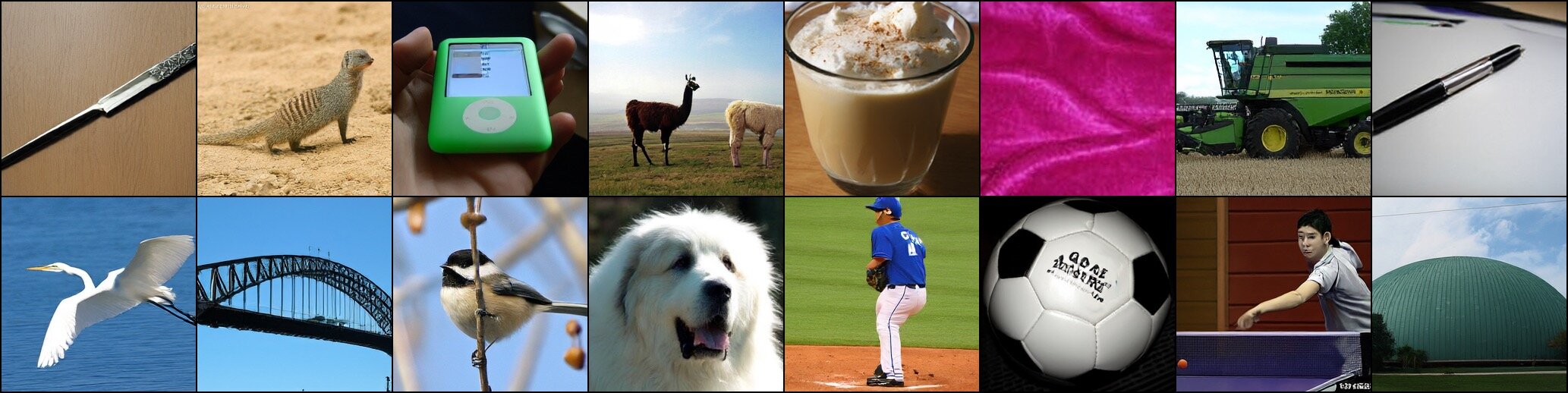}
\caption{Uncurated samples from DiT-XL/2 trained for 1M steps on top FluxAE + \regshortname (bottom) on class-conditional ImageNet $256 \times 256$ for random classes. During inference, we used 256 steps with the guidance scale of 3.0.}
\label{fig:ap:extra-samples-ditxl-flux-1m}
\end{figure*}

\begin{figure*}
\centering
\includegraphics[width=\linewidth]{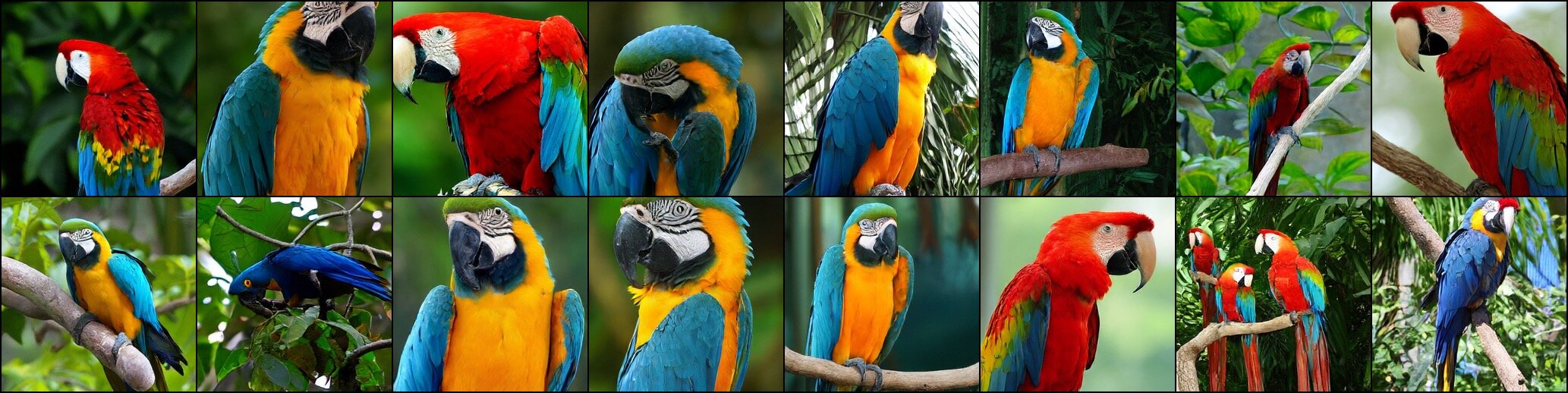}
\caption{Uncurated samples from DiT-XL/2 trained for 1M steps on top FluxAE + \regshortname (bottom) on class-conditional ImageNet $256 \times 256$ for class 88. During inference, we used 256 steps with the guidance scale of 3.0.}
\label{fig:ap:extra-samples-ditxl-flux-1m-class-88}
\end{figure*}

\begin{figure*}
\centering
\includegraphics[width=\linewidth]{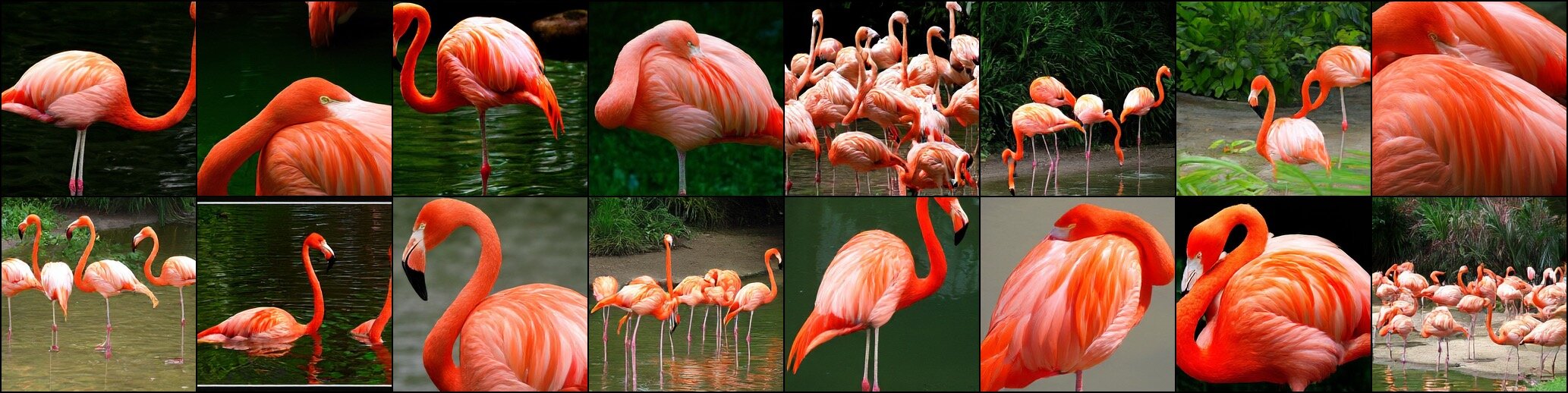}
\caption{Uncurated samples from DiT-XL/2 trained for 1M steps on top FluxAE + \regshortname (bottom) on class-conditional ImageNet $256 \times 256$ for class 130. During inference, we used 256 steps with the guidance scale of 3.0.}
\label{fig:ap:extra-samples-ditxl-flux-1m-class-130}
\end{figure*}

\begin{figure*}
\centering
\includegraphics[width=\linewidth]{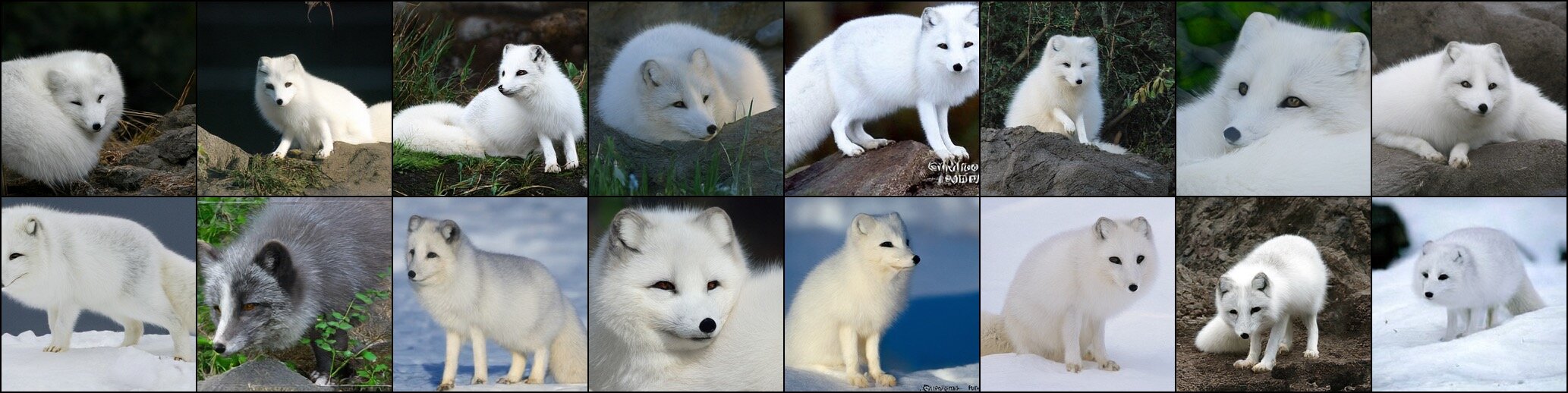}
\caption{Uncurated samples from DiT-XL/2 trained for 1M steps on top FluxAE + \regshortname (bottom) on class-conditional ImageNet $256 \times 256$ for class 279. During inference, we used 256 steps with the guidance scale of 3.0.}
\label{fig:ap:extra-samples-ditxl-flux-1m-class-279}
\end{figure*}

\begin{figure*}
\centering
\includegraphics[width=\linewidth]{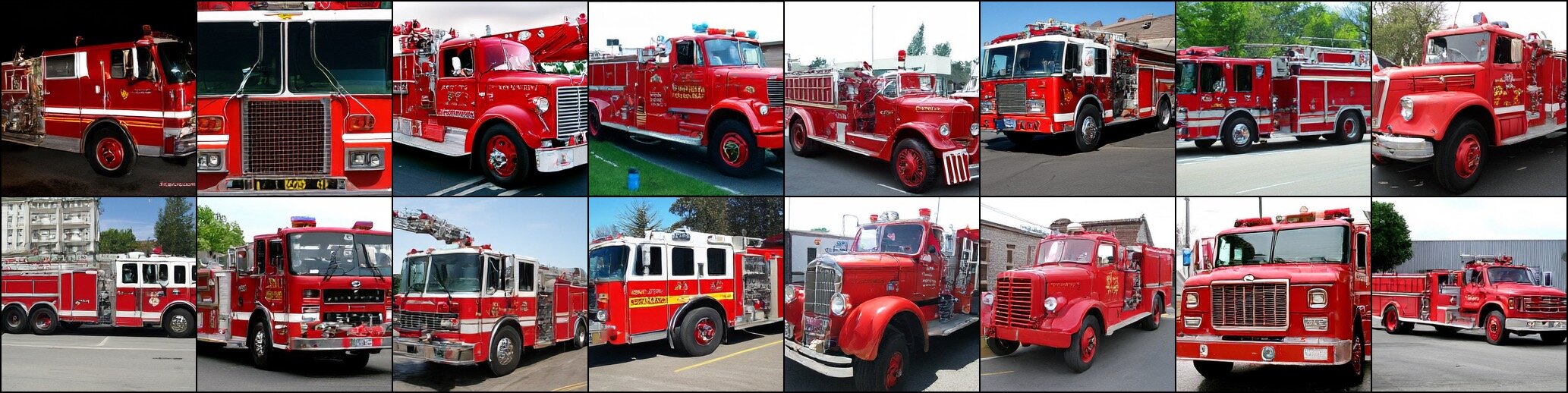}
\caption{Uncurated samples from DiT-XL/2 trained for 1M steps on top FluxAE + \regshortname (bottom) on class-conditional ImageNet $256 \times 256$ for class 555. During inference, we used 256 steps with the guidance scale of 3.0.}
\label{fig:ap:extra-samples-ditxl-flux-1m-class-555}
\end{figure*}

\begin{figure*}
\centering
\includegraphics[width=\linewidth]{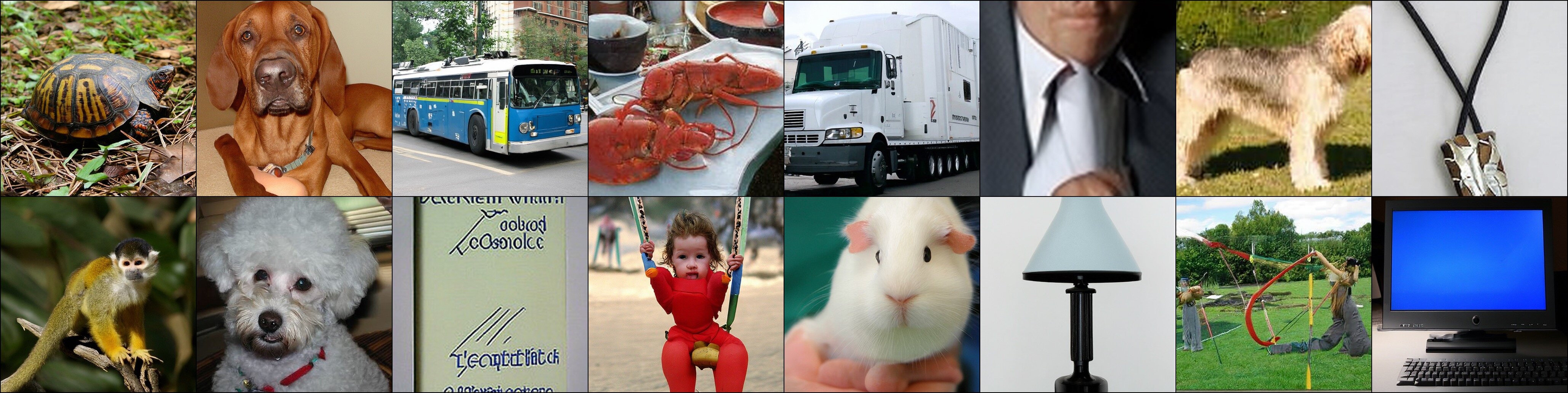}
\includegraphics[width=\linewidth]{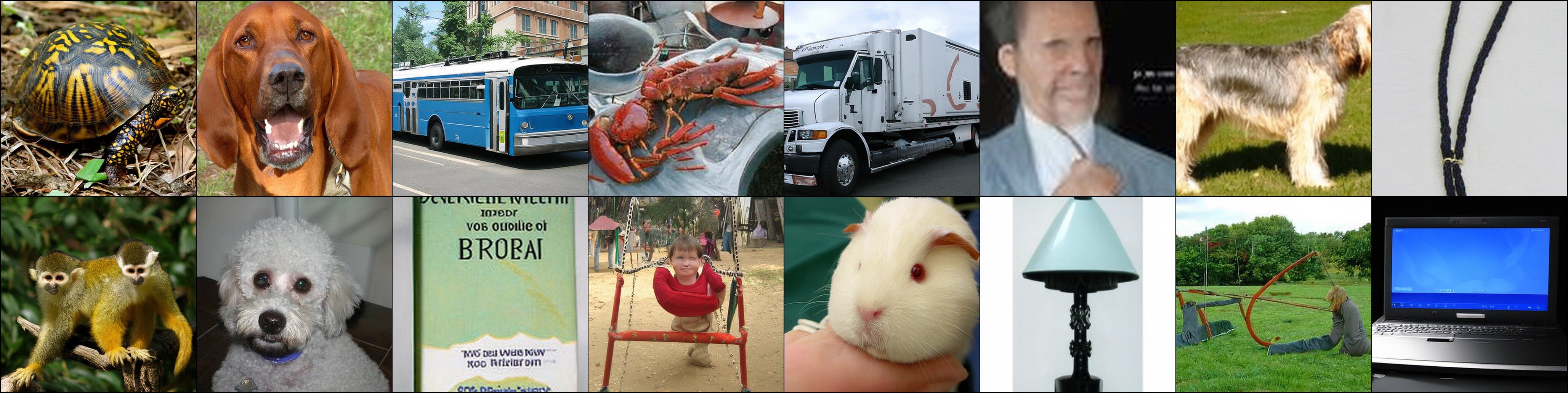}
\includegraphics[width=\linewidth]{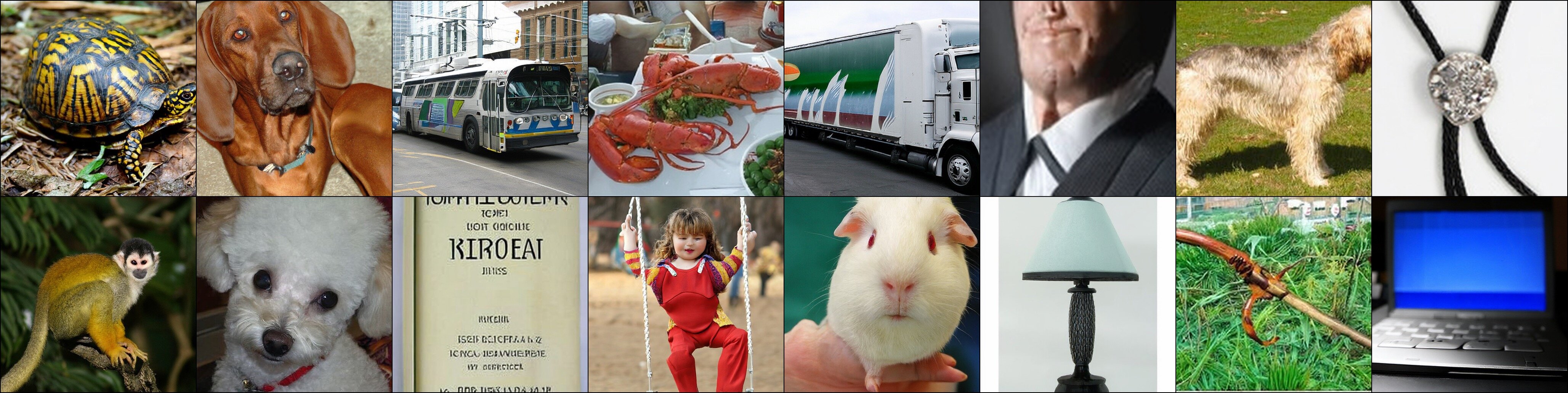}
\caption{Uncurated samples from DiT-XL/2 trained for 400K steps on top FluxAE + \regshortname (bottom) on class-conditional ImageNet $512 \times 512$ for random classes. During inference, we used 256 steps with the guidance scale of 3.0.}
\label{fig:ap:extra-samples-ditxl-flux-r512}
\end{figure*}

\begin{figure*}
\centering
\includegraphics[width=\linewidth]{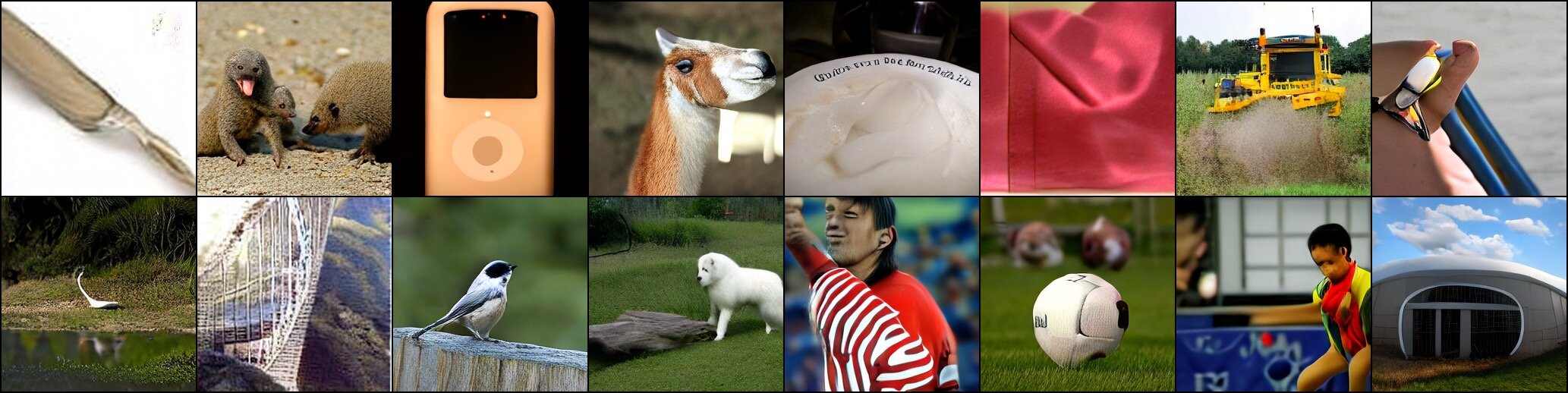}
\includegraphics[width=\linewidth]{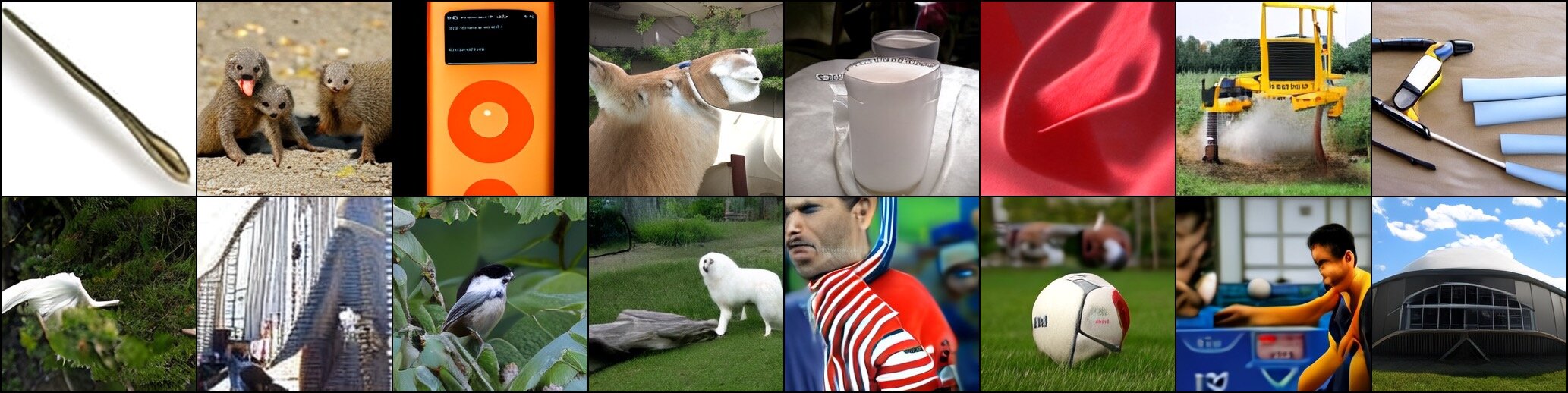}
\includegraphics[width=\linewidth]{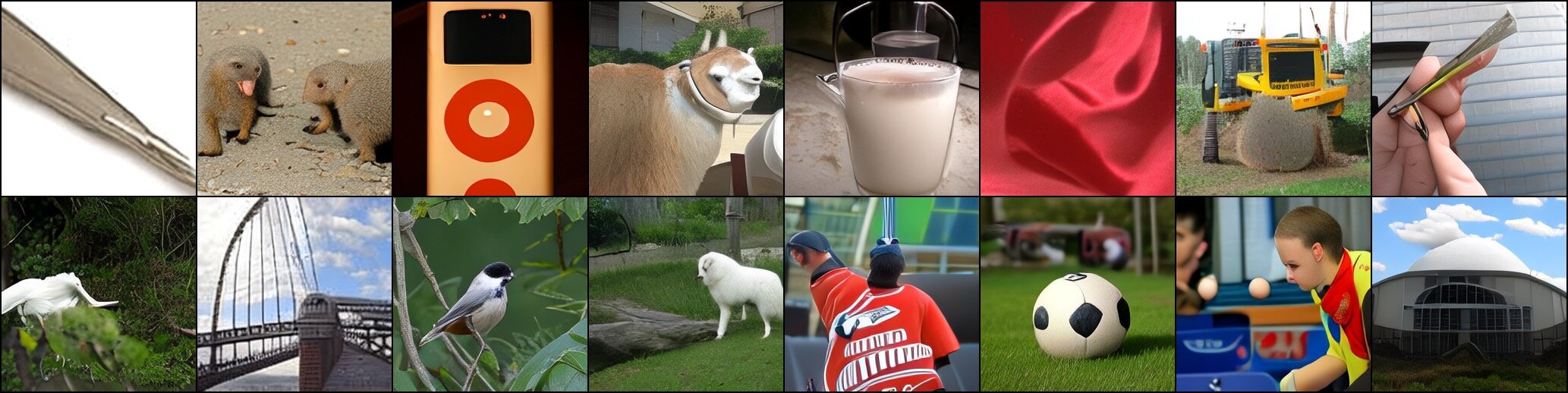}
\caption{Uncurated samples from DiT-B/1 for \cmsaei (top), \cmsaei + FT (middle) and \cmsaei + \regshortname (bottom) on class-conditional ImageNet $256 \times 256$. During inference, we used 256 steps with the guidance scale of 1.5.}
\label{fig:ap:extra-samples-ditb-cmsaei}
\end{figure*}

\begin{figure*}
\centering
\includegraphics[width=\linewidth]{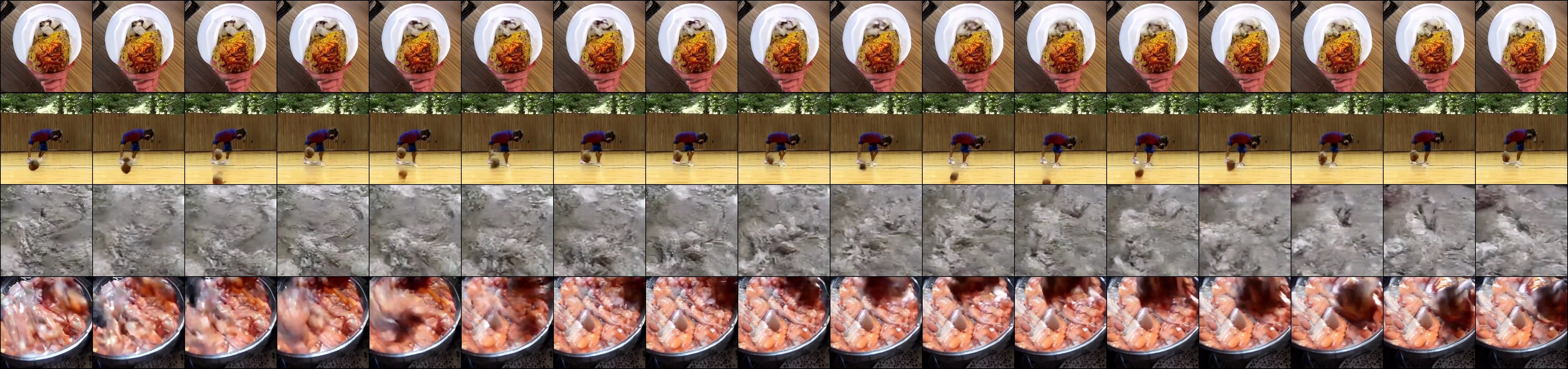}
\includegraphics[width=\linewidth]{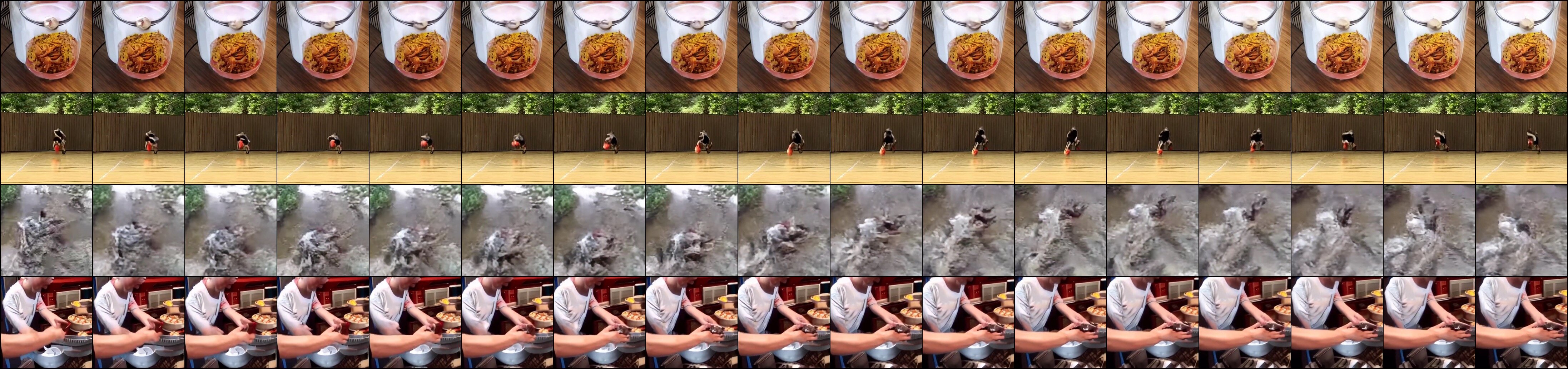}
\includegraphics[width=\linewidth]{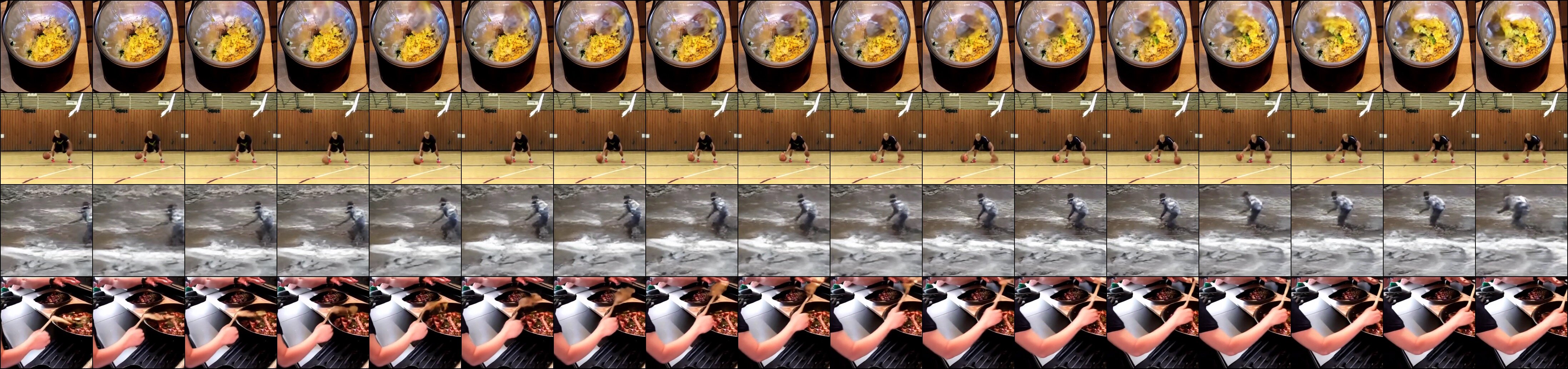}
\caption{Uncurated samples from DiT-XL/2 for \cvaefull (top), \cvaefull + FT (middle) and \cvaefull + \regshortname (bottom) on class-conditional Kinetics $17 \times 256 \times 256$. During inference, we used 256 steps with the guidance scale of 3.0.}
\label{fig:ap:extra-samples-ditxl-cvae}
\end{figure*}

\begin{figure*}
\centering
\includegraphics[width=\linewidth]{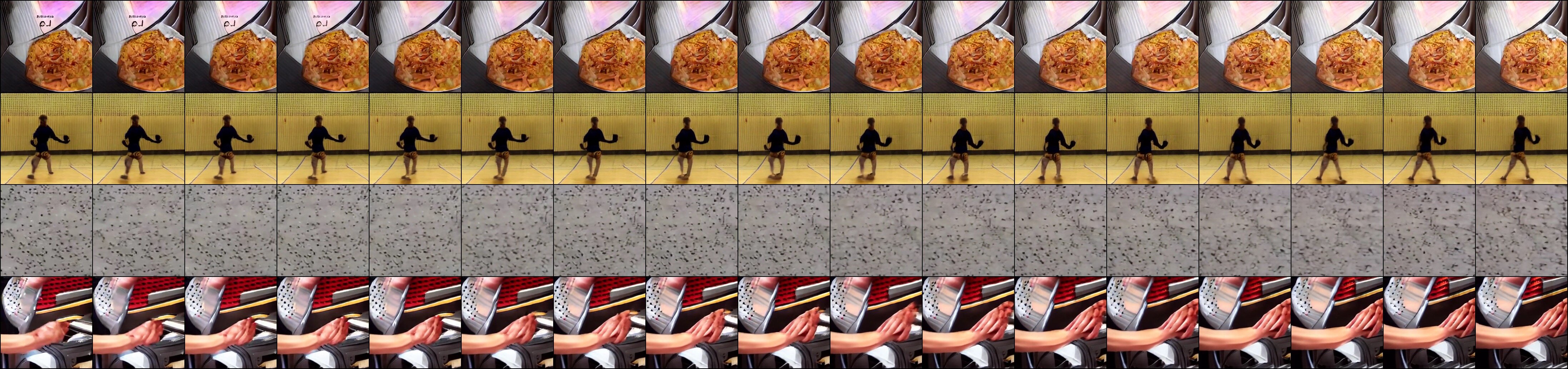}
\includegraphics[width=\linewidth]{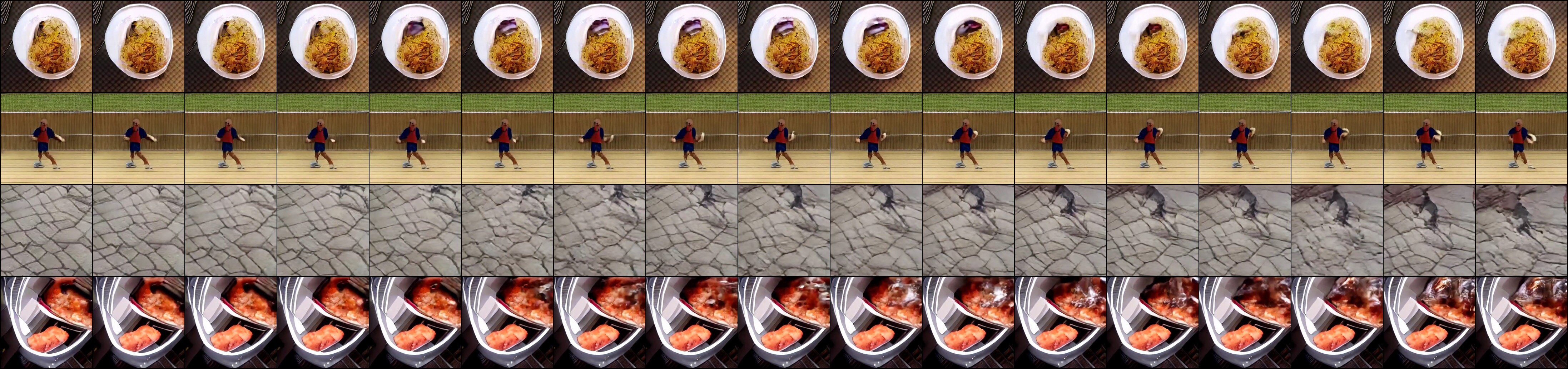}
\includegraphics[width=\linewidth]{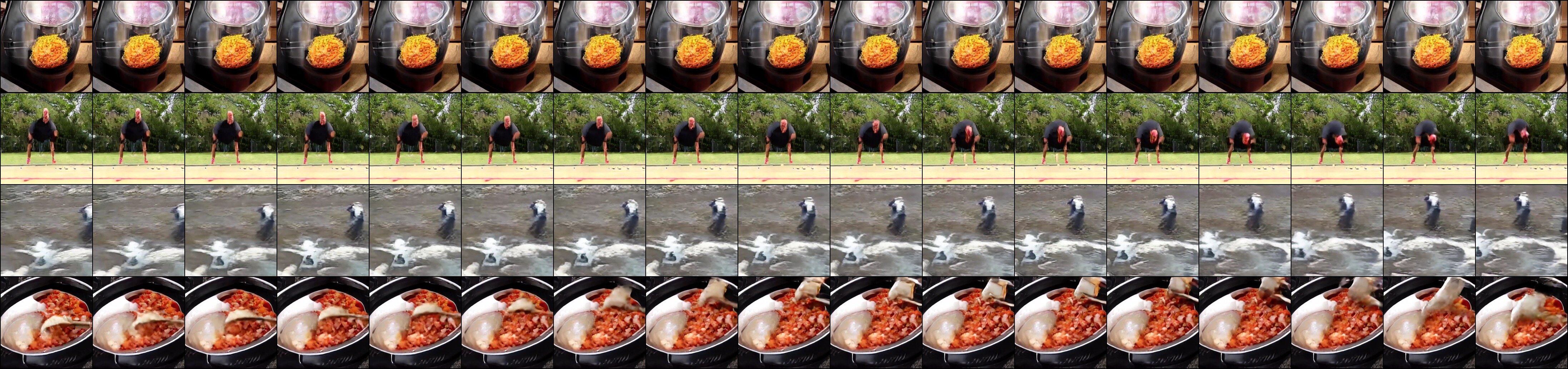}
\caption{Uncurated samples from DiT-B/2 for \cvaefull (top), \cvaefull + FT (middle) and \cvaefull + \regshortname (bottom) on class-conditional Kinetics $17 \times 256 \times 256$. During inference, we used 256 steps with the guidance scale of 3.0.}
\label{fig:ap:extra-samples-ditb-cvae}
\end{figure*}

\begin{figure*}
\centering
\includegraphics[width=\linewidth]{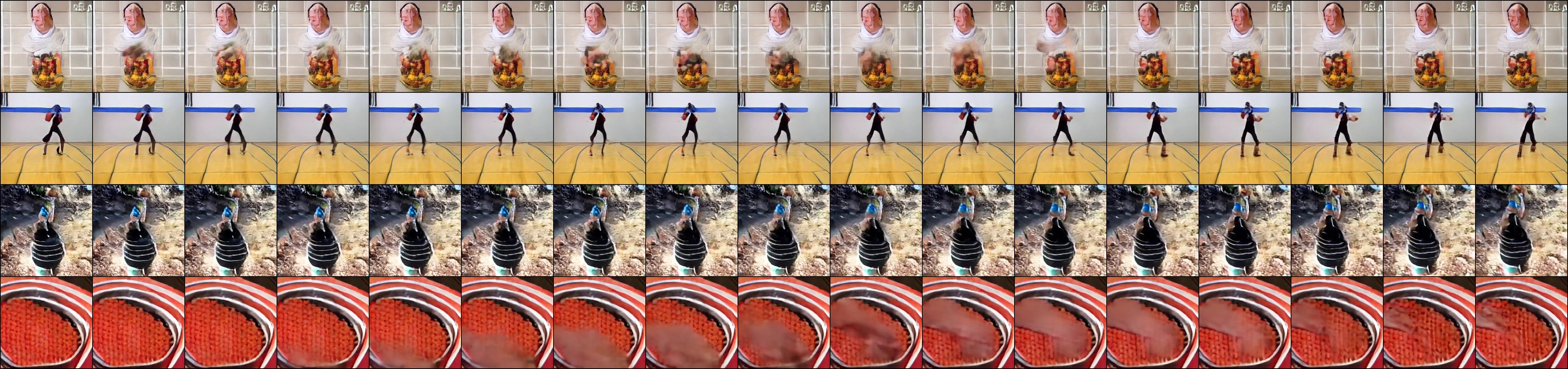}
\includegraphics[width=\linewidth]{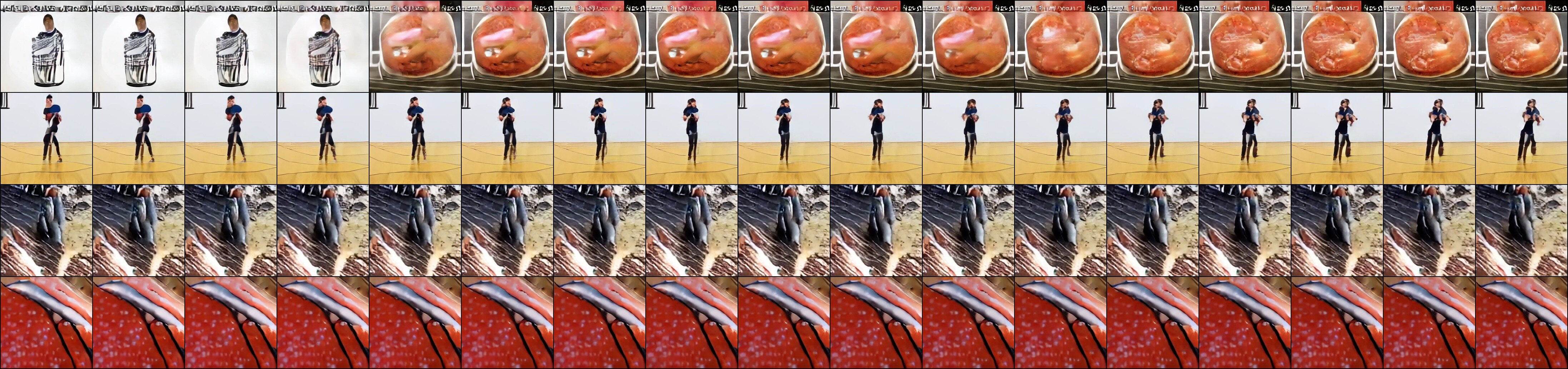}
\includegraphics[width=\linewidth]{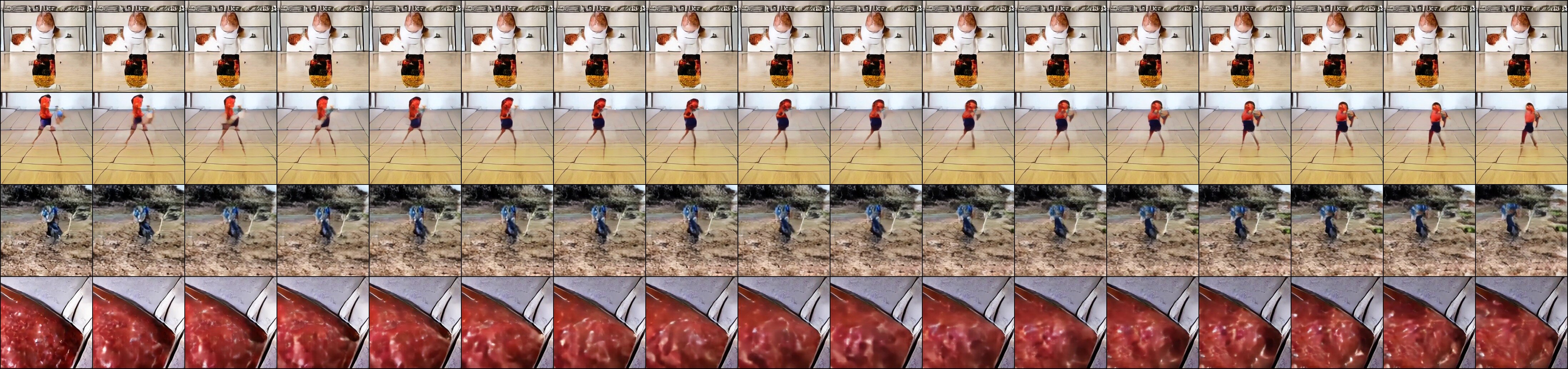}
\caption{Uncurated samples from DiT-B/1 for \ltxae (top), \ltxae + FT (middle) and \ltxae + \regshortname (bottom) on class-conditional Kinetics $17 \times 256 \times 256$. During inference, we used 256 steps with the guidance scale of 3.0.}
\label{fig:ap:extra-samples-ditb-ltxae}
\end{figure*}

%% file: supp/5-failed-experiments.tex
\section{Failed experiments}
\label{ap:failed-experiments}

Over the course of the project, we explored several other ideas to regularize the latent space of autoencoders.
While they did not pan out that well, we still discuss them in this section to spur potential future exploration.

\emph{MiniLDM-regulzation training}. LSGM~\cite{LSGM} showed that the correct objective of LDM training optimizes both autoencoder and LDM at the same time:

\begin{equation}
\begin{aligned}
\mathcal{L}(\mathbf{x}, \phi, \theta, \psi) 
&= \mathbb{E}_{q_\phi(\mathbf{z}_0|\mathbf{x})}\left[-\log p_\psi(\mathbf{x}|\mathbf{z}_0)\right] + \mathrm{KL}(q_\phi(\mathbf{z}_0|\mathbf{x})\|p_\theta(\mathbf{z}_0)) \\
&= \underbrace{\mathbb{E}_{q_\phi(\mathbf{z}_0|\mathbf{x})}\left[-\log p_\psi(\mathbf{x}|\mathbf{z}_0)\right]}_{\text{reconstruction term}} + \underbrace{\mathbb{E}_{q_\phi(\mathbf{z}_0|\mathbf{x})}\left[\log q_\phi(\mathbf{z}_0|\mathbf{x})\right]}_{\text{negative encoder entropy}} + \underbrace{\mathbb{E}_{q_\phi(\mathbf{z}_0|\mathbf{x})}\left[-\log p_\theta(\mathbf{z}_0)\right]}_{\text{cross entropy}},
\end{aligned}
\end{equation}
where $\mathcal{L}(\mathbf{x}, \phi, \theta, \psi)$ is the ELBO objective, $q_\phi, p_\psi$ are encoder/decoder, and $p_\theta$ is the LDM (or any other latent generative model).
This motivated us to explore training a small LDM model together with the autoencoder expecting that it would make its latent space more diffusable. However, even with extensive hyperparameter search, it was not outperforming the vanilla baseline.

\emph{Lipszhitz regularization}.
\citet{LFM} derived the following upper bound on the Wasserstein distance $\mathcal{W}_2^2(p_0,\hat{p}_0)$ between the ground truth $p_0(\mathbf{x})$ and recovered $\hat p_0(\mathbf{x})$ distributions, when the distribution is learned in the latent space of an autoencoder with encoder $f_\psi$ and decoder $g_\tau$:
\begin{equation}
\mathcal{W}_2^2(p_0,\hat{p}_0) \leq \|\Delta_{f_{\phi}, g_{\tau}}(\mathbf{x})\|^2 
+ L_{g_{\tau}}^2 e^{1+2\hat{L}} 
\int_0^1\int_{\mathbb{R}^{d/h}} \| v_t(\mathbf{z}_t, t) - \hat{v}_t(\mathbf{z}_t, t) \|^2 q_t^\phi \, d\mathbf{z} dt,
\end{equation}
where $v_t$ and $\hat{v}_t$ are the ground-truth and recovered velocity estimators in the rectified flow framework~\cite{RecFlow}, $L_{g_\tau}$ is the Lipschitz constant of the decoder and $\hat{L}$ is the Lipschitz constant of the learned velocity estimator $\hat{v}_t$.
This upper bound inspired us to minimize the Lipschitz constant of the decoder.
To do this, we used $R_1$-regularization~\cite{R1_reg} from the GAN literature~\cite{GANs}.
This was yielding promising initial results and working almost on par with the scale equivariance regularization, but second-order differentiation for $R_1$ regularization was entailing much slower training speed and engineering struggles (it has poor compatibility with FSDP).
This is why we proceeded with scale equivariance which was performing slightly better and much simpler conceptually.

\emph{Temporal scale equivariance}. We briefly explored temporal scale-equivariance regularization, but surprisingly, it was not leading to improved results. We hypothesize that the temporal domain is of different nature compared to the spatial one since the temporal high-frequency components are more noticeable by human eye than the low-frequency ones.